\documentclass[10pt,twocolumn,letterpaper]{article}

\usepackage[pagenumbers]{cvpr} %

\usepackage[dvipsnames]{xcolor}

\definecolor{cvprblue}{rgb}{0.21,0.49,0.74}

\usepackage[pagebackref,breaklinks,colorlinks,citecolor=cvprblue]{hyperref}

\usepackage{ifthen}
\usepackage{indentfirst}  %
\usepackage{xspace}  %
\usepackage{float}
\usepackage[frozencache=true,cachedir=minted-cache]{minted}   %

\usepackage{listings}

\usepackage[accsupp]{axessibility} %

\definecolor{cvprblue}{rgb}{0.21,0.49,0.74}

\title{Carve3D: Improving Multi-view Reconstruction Consistency \\
for Diffusion Models with RL Finetuning}

\author{Desai Xie\textsuperscript{† 1, 2} \qquad Jiahao Li\textsuperscript{† 1, 3} \qquad Hao Tan\textsuperscript{1} \qquad Xin Sun\textsuperscript{1} \qquad Zhixin Shu\textsuperscript{1}\\
Yi Zhou\textsuperscript{1} \qquad Sai Bi\textsuperscript{1} \qquad S\"oren Pirk\textsuperscript{4} \qquad Arie E. Kaufman\textsuperscript{2}\\
\\
\textsuperscript{1}Adobe Research \qquad \textsuperscript{2}Stony Brook University \qquad \textsuperscript{3}TTIC \qquad \textsuperscript{4}Kiel University
}

\newcommand\blfootnote[1]{%
  \begingroup
  \renewcommand\thefootnote{}\footnote{#1}%
  \addtocounter{footnote}{-1}%
  \endgroup
}

\begin{document}
\definecolor{TodoColor}{rgb}{0,0.5,0} 

\ifthenelse{\equal{1}{0}}  %
{
    \definecolor{TODOColor}{rgb}{0.976, 0.282, 0.235}
    \newcommand{\todos}[1]{{\color{TODOColor} \textbf{TODO}: #1}}
    
    \definecolor{DesaiColor}{rgb}{0.17, 0.8, 0.8}
    \newcommand{\desai}[1]{{\color{DesaiColor} \textbf{Desai}: #1}}
    
    \definecolor{JiahaoColor}{rgb}{0.486, 0.749, 0.482}
    \newcommand{\jiahao}[1]{{\color{JiahaoColor} \textbf{Jiahao}: #1}}
    
    \definecolor{HaoColor}{rgb}{0, 0.5, 0.88}
    \newcommand{\hao}[1]{{\color{HaoColor} \textbf{Hao}: #1}}
    
    \definecolor{XinColor}{rgb}{0.976, 0.725, 0.435}
    \newcommand{\xin}[1]{{\color{XinColor} \textbf{Xin}: #1}}
    
    \definecolor{ZhixinColor}{rgb}{0.729, 0.482, 0.976}
    \newcommand{\zhixin}[1]{{\color{ZhixinColor} \textbf{Zhixin}: #1}}
    
    \definecolor{YiColor}{rgb}{0.729, 0.478, 0.627}
    \newcommand{\yi}[1]{{\color{YiColor} \textbf{Yi}: #1}}
    
    \definecolor{SaiColor}{rgb}{0.976, 0.482, 0.725}
    \newcommand{\sai}[1]{{\color{SaiColor} \textbf{Sai}: #1}}
    
    \definecolor{SorenColor}{rgb}{0.901, 0.839, 0.349}
    \newcommand{\soren}[1]{{\color{SorenColor} \textbf{Soren}: #1}}
    
    \definecolor{ArieColor}{rgb}{0.4157, 0.3529, 0.8039}
    \newcommand{\arie}[1]{{\color{ArieColor} \textbf{Arie}: #1}}
}
{
    \newcommand{\desai}[1]{}
    \newcommand{\jiahao}[1]{}
    \newcommand{\hao}[1]{}
    \newcommand{\xin}[1]{}
    \newcommand{\zhixin}[1]{}
    \newcommand{\yi}[1]{}
    \newcommand{\sai}[1]{}
    \newcommand{\soren}[1]{}
    \newcommand{\arie}[1]{}
}

\newcommand{\secref}[1]{Section~\ref{#1}}
\newcommand{\figref}[1]{Figure~\ref{#1}}
\newcommand{\eqnref}[1]{Equation~\ref{#1}}
\newcommand{\tabref}[1]{Table~\ref{#1}}

\newcommand{\ignore}[1]{}
\newcommand\numberthis{\addtocounter{equation}{1}\tag{\theequation}}

\def\methodname{Carve3D\xspace}
\def\metricname{MRC\xspace}
\def\suppl{Supplementary Material\xspace}
\def\appx{Appendix\xspace}
\maketitle
\begin{figure}
    \centering
    \includegraphics[width=0.49\textwidth]{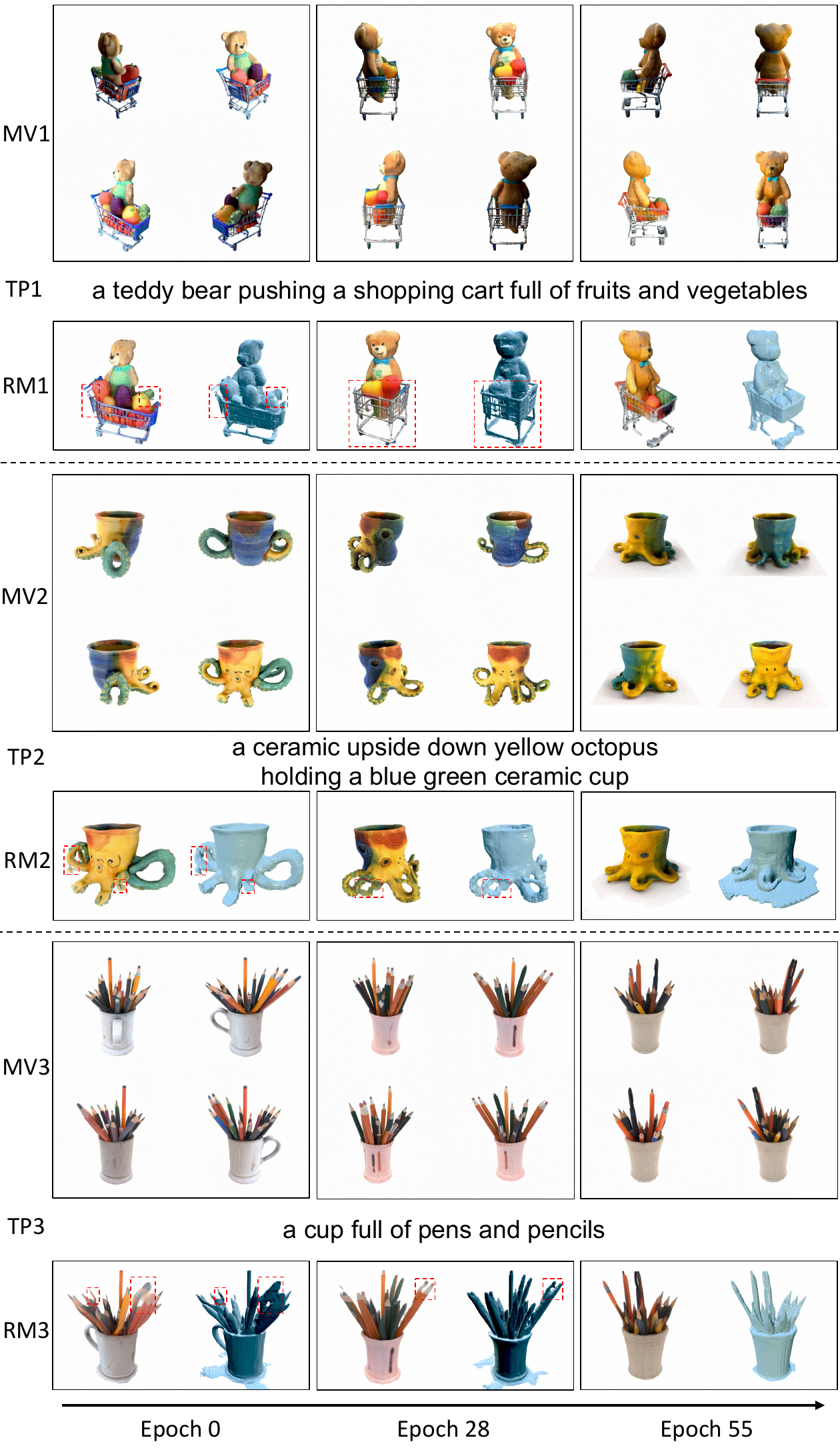}
    \vspace{-6mm}
    \caption{
        Our \methodname algorithm steadily improves the 3D consistency of a multi-view diffusion model and the resulting quality of the NeRF and the mesh, without sacrificing its image-prompt alignment, texture details, or realism. 
        Here, we show 3 testing-set results (in 3 rows, numbered as 1-3, separated by dotted lines) from the finetuning process (epoch 0, 28, and 55 in 3 columns). 
        Each row includes the generated multi-view images (denoted as MV), the reconstructed NeRF and extracted mesh (denoted as RM) and the text prompt (denoted as TP).
        The inconsistencies in the multi-view images, e.g. the facing direction of the shopping cart, the position of the octopus arms, and the position of the pencils, lead to artifacts in the NeRF and the mesh (highlighted in red).
        \vspace{-5mm}
    }
    \label{fig:teaser}
\end{figure}

\begin{abstract}

Multi-view diffusion models, obtained by applying Supervised Finetuning (SFT) to text-to-image diffusion models, have driven recent breakthroughs in text-to-3D research.
However, due to the limited size and quality of existing 3D datasets, they still suffer from multi-view inconsistencies and Neural Radiance Field (NeRF) reconstruction artifacts.
We argue that multi-view diffusion models can benefit from further Reinforcement Learning Finetuning (RLFT), which allows models to learn from the data generated by themselves and improve beyond their dataset limitations during SFT.
To this end, we introduce Carve3D, an improved RLFT algorithm coupled with a novel Multi-view Reconstruction Consistency (MRC) metric, to enhance the consistency of multi-view diffusion models. 
To measure the MRC metric on a set of multi-view images, we compare them with their corresponding NeRF renderings at the same camera viewpoints. 
The resulting model, which we denote as Carve3DM, demonstrates superior multi-view consistency and NeRF reconstruction quality than existing models. 
Our results suggest that pairing SFT with Carve3D's RLFT is essential for developing multi-view-consistent diffusion models, mirroring the standard Large Language Model (LLM) alignment pipeline. 
Our code, training and testing data, and video results are available at: \url{https://desaixie.github.io/carve-3d}.

\end{abstract}    
\blfootnote{†This work is done while the author is an intern at Adobe Research.}
\section{Introduction}
\label{sec:intro}
Recently, significant progress has been made in generating 3D models from text prompts. Images generated by 2D diffusion models  ~\cite{rombach2022sd,poole2022dreamfusion,wang2022sjc,chen2023GSGEN,GaussianDreamer,tang2023dreamgaussian} can be lifted to 3D representations. Numerous methods~\cite{shi2023MVDream,li2023instant3d,liu2023syncdreamer,zhao2023efficientdreamer} have demonstrated that a set of multi-view images is adequate for generating diverse and detailed 3D models, effectively mitigating the multi-face (Janus) problem.
Ensuring the 3D consistency across these multi-view images is crucial for 3D generation, as inconsistencies can inevitably introduce artifacts, such as broken geometries, blurring, or floaters, in the Neural Radiance Field (NeRF) reconstruction.
However, the lack of an established multi-view consistency metric has led researchers to rely on qualitative inspections, which are both inefficient and unreliable.%

Existing multi-view diffusion models~\cite{shi2023MVDream,li2023instant3d,liu2023syncdreamer,zhao2023efficientdreamer,liu2023zero1to3} primarily utilize Supervised Finetuning (SFT) with multi-view datasets derived from 3D datasets~\cite{deitke2022objaverse, deitke2023objaversexl}.
While SFT can achieve some degree of multi-view consistency, it presents a dilemma: prolonged SFT enhances multi-view consistency but also induces a distribution shift towards the 3D dataset, which has limited size and quality; thus, the distribution shift diminishes diversity, texture details, and realism of the generated results~\cite{li2023instant3d}.
Such dilemma has been observed in Large Language Model (LLM) research.
While SFT can shift the output distribution of pre-trained LLMs to follow instructions, the bias from the instruction dataset also introduces hallucination~\cite{schulman2023rlhf}, preventing longer SFT. 
InstructGPT~\cite{ouyang2022InstructGPT}, the paper behind ChatGPT 3.5~\cite{openai_chatgpt}, introduces Reinforcement Learning finetuning (RLFT) to further align the SFT model without causing additional distribution shift.
Drawing an analogy between instruction-finetuned LLMs and multi-view diffusion models, RLFT emerges as an essential step following the SFT stage.
By adopting RLFT, we aim to enhance the consistency of multi-view diffusion models without introducing the bias from a SFT dataset.

\begin{figure*}
    \centering
    \includegraphics[width=0.95\textwidth]{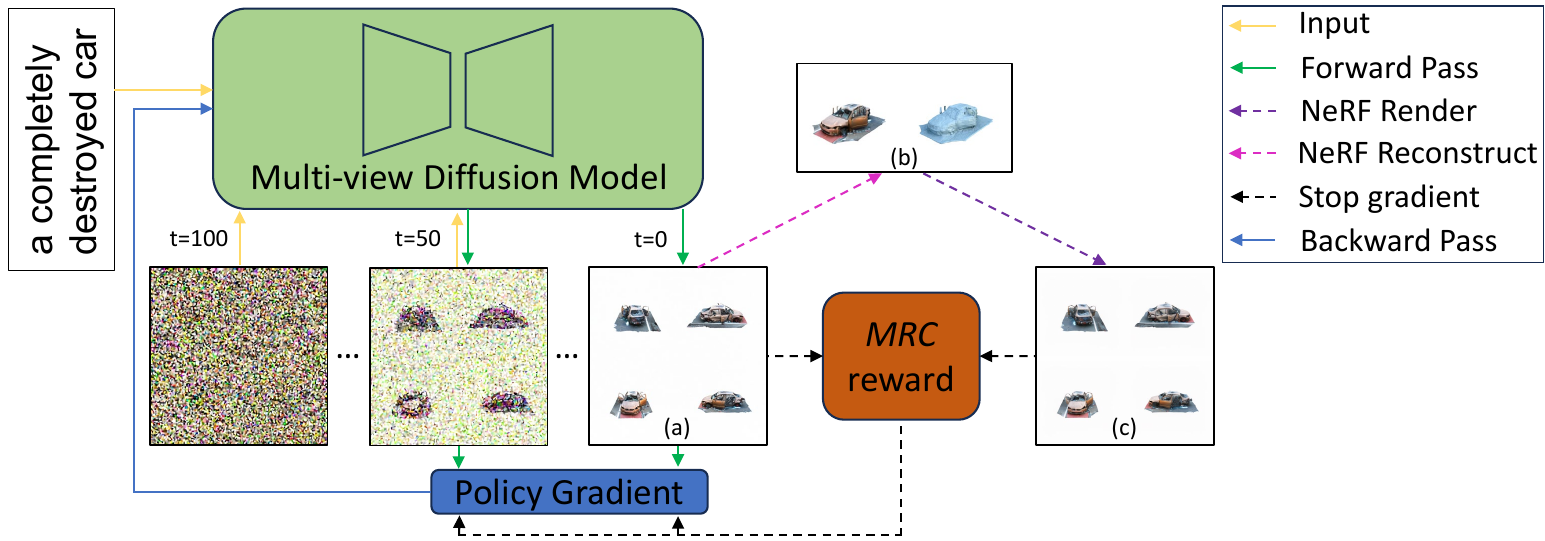}
    \caption{
        Overview of Carve3D. 
        Given a prompt sampled from our curated prompt set and a initial noisy image, we iteratively denoise the image using the UNet. 
        The final, clean image contains four multi-view images tiled in a 2-by-2 grid.
        \metricname reward is computed by comparing (a) the generated multi-view images with (c) the corresponding multi-view images rendered at the same camera viewpoints from (b) the reconstructed NeRF.
        Then, we train the model with policy gradient loss function, where the loss is derived from the reward and log probabilities of the UNet's predictions, accumulated over all denoising timesteps.
        By using only a set of training text prompts, our RLFT algorithm finetunes the diffusion model by evaluating its own generated outputs, without relying on ground truth multi-view images.
        \vspace{-3mm}
    }
    \label{fig:overview}
\end{figure*}

We introduce \methodname, an enhanced RLFT algorithm paired with a novel Multi-view Reconstruction Consistency (MRC) metric, to improve the consistency of multi-view diffusion models.
\cref{fig:teaser,fig:overview} shows the capability and overview of Carve3D.

Our MRC metric compares the output multi-view images from a diffusion model with images rendered from the reconstructed NeRF at identical camera viewpoints.
We use the sparse-view Large Reconstruction Model (LRM)~\cite{hong2023lrm,li2023instant3d} to achieve fast, feed-forward NeRF reconstruction from a few multi-view images.
To quantify image similarity, we adopt LPIPS~\cite{zhang2018lpips} as it is more effective and robust for \metricname.
We further normalize LPIPS with respect to the bounding boxes of foreground objects to prevent trivial reward hacking through size reduction of the foreground object.
To validate the reliability of \metricname, we conduct extensive experiments with controlled inconsistency levels; 
starting from a set of perfectly consistent multi-view images rendered from a 3D asset~\cite{deitke2022objaverse}, we manually introduce distortion to one of the views to create inconsistency.
Our MRC metric provides robust evaluation of consistency of multi-view images, offers a valuable tool for assessing current multi-view generation methods and guiding future developments in the field.

With \metricname, we employ RLFT for multi-view diffusion models.
In the RLFT process, we use a set of curated, creative text prompts to repeatedly generate diverse multi-view images with random initial noises and use their MRC reward to update the diffusion model (\cref{fig:overview}).
Such diversity- and quality-preserving finetuning cannot be achieved with SFT, as it is infeasibly expensive to create a dataset of diverse ground-truth multi-view images for these prompts.
We make the following improvements to the RLFT algorithm~\cite{black2023DDPO}.
In addressing the common training instability issue in RL, we opt for a purely on-policy policy gradient algorithm~\cite{williams1992REINFORCE}, diverging from the widely adopted, partially on-policy PPO~\cite{schulman2017ppo} algorithm.
We incorporate KL divergence regularization~\cite{fan2023dpok, ouyang2022InstructGPT} to maintain proximity to the base model and prevent distribution shift.
Moreover, we scale up the amount of compute to achieve optimal rewards by applying the scaling laws for diffusion model RLFT, identified from extensive experiments -- a topic that has not yet been extensively covered in existing studies~\cite{black2023DDPO,fan2023dpok}. 

By adopting our Carve3D RLFT algorithm on Instant3D-10K~\cite{li2023instant3d}, a multi-view diffusion model supervised finetuned from SDXL~\cite{podell2023sdxl}, we obtain the Carve3D Model (Carve3DM).
Through quantitative, qualitative experiments and a user study, we demonstrate that Carve3DM: 
(1) achieves improved multi-view consistency and NeRF reconstruction quality over Instant3D-10K, -20K, and -100K models, and 
(2) maintains similar prompt alignment, diversity, and realistic details as the base Instant3D-10K, preventing the degradation in Instant3D-20K and -100K.
Our results indicate that pairing SFT with Carve3D's RLFT is essential for developing multi-view consistent diffusion models.
In addition, we extend our MRC evaluation to existing models, revealing the universal presence of multi-view inconsistency when relying solely on SFT.
Our work is the first application of RLFT to text-to-3D, especially on a 2.6B-parameter denoising UNet from SDXL~\cite{podell2023sdxl}. 
By releasing our code, training and testing data, we hope this work will bolster the RLFT and alignment research in the computer vision community.

\section{Related Works}
\label{sec:bg:textto3D}

Neural Radiance Field (NeRF) is a neural representation of 3D assets~\cite{mildenhall2020nerf,Chen2022tensorf,mueller2022instantngp}.
It infers the direction-dependent radiance at arbitrary volumetric positions with neural models.
Many text-to-3D methods rely on it to produce 3D objects. 

While text-to-image diffusion models are trained on 5 billion data~\cite{schuhmann2022laion5b}, the largest public 3D dataset only contains 10 million 3D assets~\cite{deitke2022objaverse,deitke2023objaversexl} with little text annotation.
This gap in the diversity and quality of 3D data has restricted the quality of current 3D diffusion models and their ability in handling complex prompts~\cite{nichol2022pointe,jun2023ShapE}. 
To circumvent this limitation, another line of work focuses on lifting 2D images to 3D, thus leveraging the remarkable semantic understanding and high-quality generation capabilities of 2D diffusion models~\cite{rombach2022sd,podell2023sdxl}.
These methods~\cite{poole2022dreamfusion,wang2022sjc,chen2023GSGEN,GaussianDreamer,tang2023dreamgaussian} typically employ 2D diffusion models to provide supervision at the novel views for optimizing 3D objects represented as NeRF or by 3D Gaussian Splatting~\cite{kerbl2023GaussianSplatting}.
Building on this concept, multiple works~\cite{liu2023zero1to3,liu2023one2345,shi2023MVDream,li2023instant3d,liu2023syncdreamer,zhao2023efficientdreamer} have proposed generating multi-view images using a finetuned 2D diffusion model, providing a more comprehensive visual prior and preventing the multi-face (Janus) problem.
However, as the finetuning datasets of multi-view images are rendered from the same 3D dataset~\cite{deitke2022objaverse,deitke2023objaversexl}, the limited quality and diversity remains a challenge, preventing running Supervised Finetuning to convergence~\cite{li2023instant3d}.
By adopting Reinforcement Learning Finetuning (RLFT), we do not depend on ground truth multi-view images and thus optimize the model beyond the distribution of their SFT dataset.

A key challenge in utilizing multi-view images is achieving 3D consistency.
While numerous methods have attained notable multi-view consistency by supervised finetuning 2D diffusion models~\cite{shi2023MVDream,li2023instant3d,liu2023syncdreamer,zhao2023efficientdreamer,liu2023zero1to3}, their evaluation has been empirical, lacking explicit metrics. 
An approach known as 3D consistency scoring~\cite{watson2022novel} measures the consistency of output views by optimizing a NeRF trained on these views. 
Our \metricname metric improves it with sparse-view reconstruction and comparing on all input views.

3D models can be derived from either single or multi-view images by optimizing the Score Distillation Sampling (SDS) loss~\cite{poole2022dreamfusion, wang2022sjc}. 
However, the optimization process is notably time-consuming, requiring multiple hours to generate a single 3D asset.
In contrast, Large Reconstruction Model (LRM)~\cite{hong2023lrm}, trained on the extensive 3D dataset Objaverse~\cite{deitke2022objaverse}, can efficiently reconstruct NeRF models from a single image in a feed-forward manner. 
In this work, we focus exclusively on text-to-3D using feed-forward sparse-view NeRF reconstruction, specifically employing sparse-view LRM~\cite{li2023instant3d}. 
This choice is driven by its significantly faster performance compared to SDS-based optimization methods and its superior quality relative to feed-forward text-to-3D diffusion models~\cite{nichol2022pointe, jun2023ShapE}. 
We choose Instant3D~\cite{li2023instant3d} as our base multi-view diffusion model, owing to its light-weight Supervised Finetuning (SFT) that preserves the strong semantic understanding and high-quality image generation capabilities of SDXL~\cite{podell2023sdxl}, similar to the instruction finetuning stage in InstructGPT~\cite{ouyang2022InstructGPT}. 

\subsection{RLFT of LLMs and Diffusion Models}
\label{sec:suppl:related_work_rlft}
Reinforcement Learning (RL) has been widely used to finetune large pre-trained models in Natural Language Processing~\cite{ouyang2022InstructGPT, bai2022RLAIF, lee2023rlaif, bai2022helpfulharmlessRLHF} and Computer Vision~\cite{black2023DDPO,pinto2023tuning_cv_rl, fan2023dpok, prabhudesai2023alignprop, clark2023DRaFT, zhang2022hive}, due to its advantage over Supervised Finetuning (SFT).
SFT directly fits a model to the distribution of the SFT dataset containing inputs and ground-truth target data, which unavoidably causes some degree of distribution shift~\cite{schulman2023rlhf}.
On the contrary, based on an objective function and a dataset containing only inputs, Reinforcement Learning Finetuning (RLFT) optimizes a model beyond the limitation of a SFT dataset by using its own outputs and effectively mitigates distribution shift~\cite{casper2023RLHFproblems}. 

\paragraph{RLFT of LLMs}
Large Language Models (LLMs) like GPT-3~\cite{brown2020gpt3} are pre-trained on the next-word prediction task on an internet-scale corpus.
While the autoregressive pre-training is a powerful self-supervised objective that allows LLMs to extract substantial knowledge from the internet-scale unlabeled dataset, pre-trained LLMs can only perform the corresponding text completion task.
The pre-training lacks an objective that allows LLMs to respond to text prompts.
In InstructGPT~\cite{ouyang2022InstructGPT}, the paper behind ChatGPT 3.5, a two-stage finetuning solution is proposed to align GPT-3 to answer instructions according to human preferences.
In the first stage, InstructGPT employs SFT with a small dataset of hand-crafted prompt-answer pairs.
While SFT changes the model's output distribution from text completion to answering instructions, it also introduces hallucination~\cite{schulman2023rlhf}.
This is because the output distribution drifts too much towards the instruction-following dataset, and the model tries to imitate the behavior in the data and always provide plausible answers even when the model is uncertain about the answer~\cite{schulman2023rlhf}.
To address this issue, InstructGPT opts for a light-weight SFT stage and relies on RLFT in the second stage, using a human-preference reward model.
This approach provides general alignment to human values and causes minimal hallucination~\cite{schulman2023rlhf}, because RLFT does not rely on a potentially biased dataset containing fixed ground-truth answers, but instead learns the general concept of human-preference through the reward model. 
The success of InstructGPT~\cite{ouyang2022InstructGPT} and its analogy to the distribution shift problem in multi-view SFT~\cite{li2023instant3d} motivate us to pursue RLFT for 2D diffusion models.

\paragraph{RLFT of Diffusion Models}

Witnessing the success of RLFT methods in LLMs~\cite{ouyang2022InstructGPT,bai2022helpfulharmlessRLHF,bai2022RLAIF,lee2023rlaif}, recently, a few RLFT algorithms have been proposed for text-to-image diffusion models.
RWR~\cite{lee2023rwr} is the first work to bring the human feedback reward finetuning idea to diffusion models.
While RWR only finetunes stable diffusion~\cite{rombach2022sd} via a single log probability of the entire denoising process,
multi-step RLFT can be facilitated by treating the denoising process as a multi-step MDP, as demonstrated in DDPO~\cite{black2023DDPO} and DPOK~\cite{ fan2023dpok}.
Our RLFT is based on DDPO~\cite{black2023DDPO}, while 
our KL-divergence regularization is similar to DPOK~\cite{fan2023dpok} and InstructGPT~\cite{ouyang2022InstructGPT}. Furthermore, RWR, DDPO, and DPOK all finetune SD-1.5~\cite{rombach2022sd}, while we finetune a much larger diffusion model based on SDXL.
We also study training stability, a notorious challenge in both traditional RL and RLFT~\cite{zheng2023RLHFsecrets,casper2023RLHFproblems}, and scaling laws~\cite{kaplan2020scalinglaws} for RLFT.

\begin{figure*}
    \centering
    \includegraphics[width=0.85\linewidth]{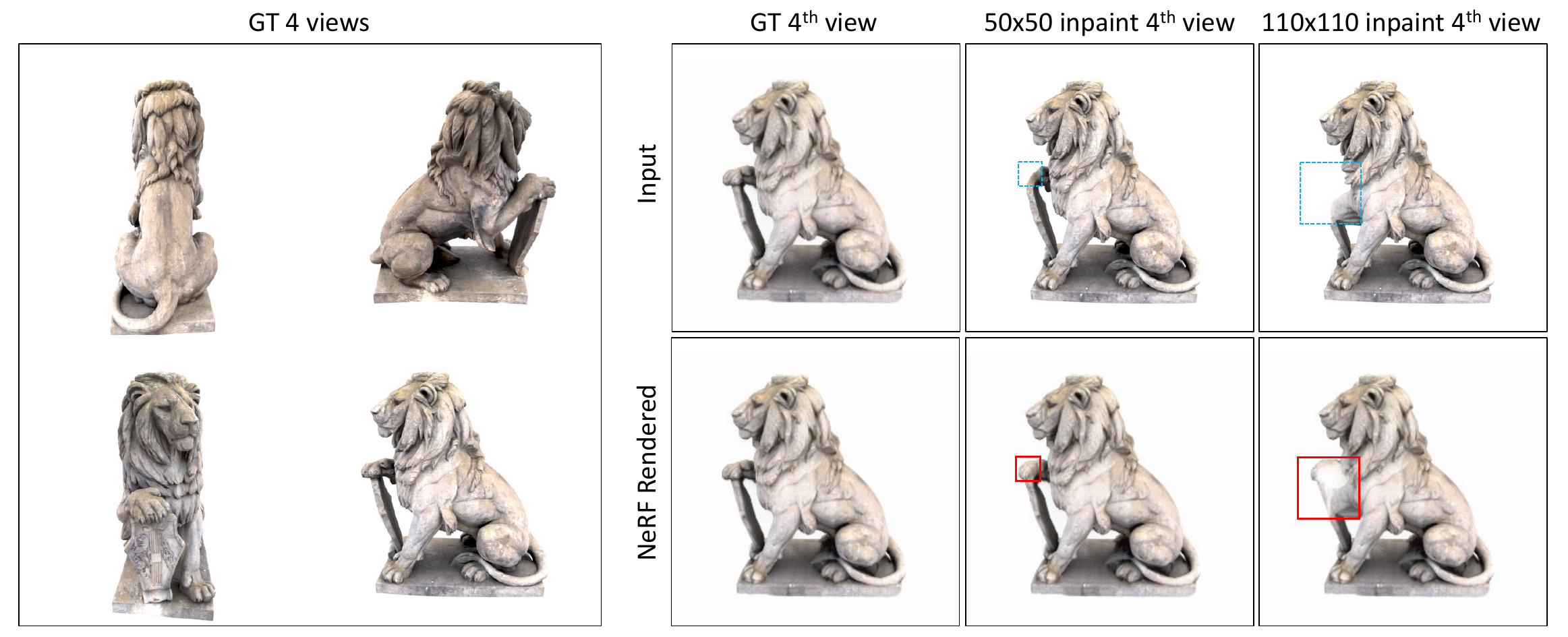}
    \caption{
        Qualitative correlation between \metricname and multi-view inconsistency with increasing intensity, introduced by inpainting with increasing mask sizes.
        Left: the four ground truth views.
        Right: the 4th view is inpainted with increasing area sizes, i.e. 0$\times$0, 50$\times$50 and 110$\times$110 pixels.
        The top row is the image after inpainting and the bottom row is the image rendered from the NeRF reconstructed with the top inpainted 4th view and the other 3 original GT views.
        We mark the inpainting area with blue and red boxes.
        Since the lion's right paw in the inpainted 4th views look different from the other three original views, its shape is broken in the NeRF and the rendered views. 
        This difference is captured in MRC's image dissimilarity metric.
        \vspace{-3mm}
}
    \label{fig:method:consistency}
    
\end{figure*}
\begin{figure}
    \centering
    \includegraphics[width=0.35\textwidth]{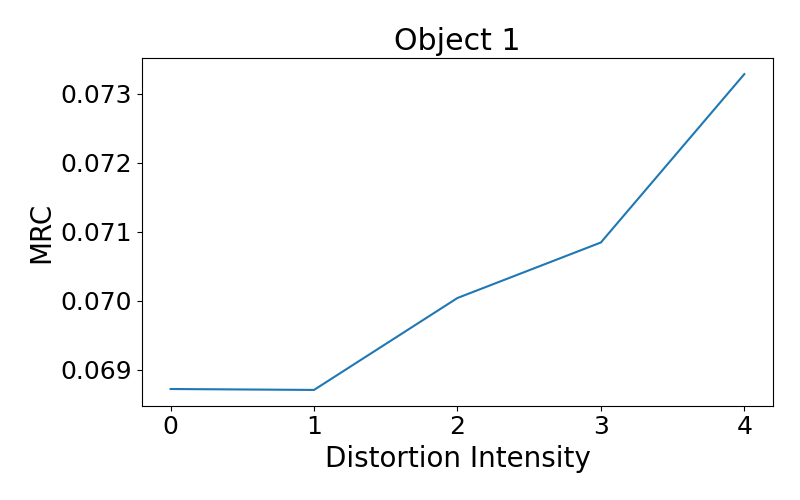}
    \vspace{-3mm}
    \caption{
    Quantitative correlation between \metricname and multi-view inconsistency with increasing intensity, for the object shown in~\Cref{fig:method:consistency}.
    As inconsistency intensity rises, \metricname also monotonically increases.
    \vspace{-3mm}
    }
    \label{fig:method:consistency-plot}
\end{figure}
\section{Multi-view Reconstruction Consistency}
\label{sec:metric}

In this section, we propose the Multi-view Reconstruction Consistency (MRC) metric, for quantitative and robust evaluation of the consistency of multi-view images, which we define to be \textit{the degree of geometry and appearance uniformity of an object across the views}.

\subsection{Evaluate Consistency via NeRF Reconstruction}
\label{sec:metric:viaNeRF}

A 3D model represented by Neural Radiance Field (NeRF) can be reconstructed from the view images of the object and their corresponding camera poses.
The quality of a NeRF notably depends on the consistency of the provided images images~\cite{mildenhall2020nerf, watson2022novel} -- inconsistent views lead to artifacts in the NeRF, which includes floaters, blurring, and broken geometry.
To address this challenge, we introduce a metric for assessing the consistency among multiple views.%

The intuition behind MRC comes from the relationship between multi-view consistency and the reconstructed NeRF.
As shown in~\cref{fig:method:consistency}, when the multi-view images are consistent, they can produce a well reconstructed NeRF, preserving almost all the visual cues from the input images;
therefore, the views rendered from the NeRF at the same camera viewpoints will look the same as the original views;
conversely, when the multi-view images are inconsistent (e.g., intentionally introduced inconsistency in~\cref{fig:method:consistency}), they will produce a NeRF with broken geometry and floater artifacts; 
thus, the NeRF rendered views will look different from the original views.
Building upon this observation, we propose the \metricname metric, defined as the image distances between the original multi-view images and the views of the reconstructed NeRF rendered at the same viewpoints, as illustrated in~\cref{fig:overview}.

\subsection{Implementation}
\label{sec:metric:implementation}
We formulate the implementation of \metricname as three parts: fast sparse-view NeRF reconstruction, measuring image distance between the input images and the rendered images, and a normalization technique for the image distance.
The pseudo code for our MRC implementation is shown in~\appx\cref{lst:mrc}. 

\paragraph{Fast Sparse-view Reconstruction}
We conduct NeRF reconstruction with sparse-view Large Reconstruction Model (LRM) proposed in ~\cite{li2023instant3d,hong2023lrm}.
Different from dense view NeRF reconstruction~\cite{mildenhall2020nerf,Chen2022tensorf,mueller2022instantngp}, sparse-view LRM reconstructs a NeRF with only $4$-$6$ view images.
Also, with its feed-forward reconstruction, it can achieve a speed two orders of magnitude faster than previous optimization-based reconstruction methods.
\metricname leverages all multi-view images for both NeRF reconstruction and 3D consistency evaluation.
Although the NeRF is reconstructed based on the visual prior of the input multi-views images, the rendering from the same views still exhibits notable differences if there is inconsistency inside the input, as shown in~\cref{fig:method:consistency}.

\paragraph{Image Distance Metric}
In~\cref{sec:metric:viaNeRF}, the consistency problem is reduced from 3D to a 2D image dissimilarity problem.
To measure the image dissimilarity between the input views and their corresponding NeRF rendered views, we utilize the perceptual image distance metric, LPIPS~\cite{zhang2018lpips}.
LPIPS exhibits smoother value changes with respect to the consistency of multi-view images compared to PSNR, SSIM, L1, and L2, as shown in~\appx\cref{fig:suppl:metricexp:inpaint}.
Such smoothness is derived from the non-pixel-aligned computation in LPIPS, as opposed to the other image distance metrics that are more pixel-aligned. 
Also, the smoothness is a crucial aspect for MRC to serve as the reward function in RLFT, because non-smooth, high-variance reward functions makes the RLFT training more challenging.

\paragraph{Bounding-box Normalization} 
Current multi-view diffusion models~\cite{shi2023MVDream,li2023instant3d,liu2023syncdreamer,zhao2023efficientdreamer} target single object generation with background.
Consequently, 
if computing LPIPS on the entire image, trivially reducing the object's relative size (as illustrated in~\appx\cref{fig:suppl:bbox_norm}'s car example) can exploit \metricname, as the majority of images will be the white background.
Therefore, we propose normalizing our metric with respect to the object's size.
Specifically, we identify the smallest square bounding box of the foreground object in the input view image.
Then we crop both the input images and the rendered images with that bounding box, resize them to a fixed resolution, and evaluate the LPIPS.
This normalization effectively prevents the reward hacking of \metricname by diminishing foreground object sizes, as shown in~\appx\cref{fig:suppl:bbox_norm}.

\subsection{Metric Experiment}
\label{sec:metric:exp}
The two key objectives for introducing the \metricname metric are (1) to assess the consistency of any multi-view generative model and (2) to enable  RLFT for improving the consistency of multi-view diffusion models.
Thus, the proposed consistency metric should ideally present two respective properties: 
(1) MRC should monotonically increase as inconsistency increases;
(2) the MRC vs. inconsistency curve should be smooth.

To validate the effectiveness and robustness of \metricname, i.e. whether it satisfies the two properties, we conduct evaluation on sets of multi-view images with controlled level of inconsistency.
Starting from a set of perfectly-consistent ground truth views rendered from a 3D asset from Objaverse~\cite{deitke2022objaverse}, we manually introduce inconsistency to one image.
We select a portion of this image and inpaint it with an image-to-image diffusion model\footnote{We use Adobe Photoshop's Generative Fill~\cite{adobe_firefly} without text prompt to add inpainting distortion, which is based on a diffusion model.}.
Therefore, we get different levels of distortion on one image, determined by the size of the inpainting area, that corresponds to different levels of inconsistency of the set of images.

\cref{fig:method:consistency} shows the qualitative result on one object of our MRC metric experiment.
With increased inpainting area size, the NeRF rendered view also shows larger image difference, which is then captured by MRC's image distance metric, LPIPS.
\cref{fig:method:consistency-plot} presents the quantitative curve of the same experiment. 
\metricname indeed shows a monotonically increasing pattern as the views become more inconsistent.
As shown in~\appx\cref{fig:suppl:metricexp:inpaint}, MRC constantly exhibits monotonically increasing pattern, and it is also smoother than the other MRC variants using PSNR, SSIM, L1, and L2.
For metric experiments on other distortion types, see~\cref{sec:suppl:mrc_exp}.

\section{RLFT for Multi-view Consistency}
\label{sec:RL}

In~\cref{sec:metric}, we proposed a fast and reliable multi-view consistency metric named MRC, and in this section we describe how it can be used to finetune a multi-view diffusion model.
In this section, we present an improved Reinforcement Learning Finetuning (RLFT) algorithm for enhancing the consistency of 2D multi-view diffusion models, using the negative \metricname as the reward function (\appx\cref{lst:mrc}). 
Building upon DDPO~\cite{black2023DDPO}, we opt for its pure on-policy version over the default partially on-policy version of the policy gradient algorithm for substantially improved training stability.
To maintain proximity to the base model, we incorporate KL divergence regularization similar to ~\cite{fan2023dpok,ouyang2022InstructGPT}.
In addition, we scale up the RLFT to achieve higher rewards by studying the scaling laws~\cite{kaplan2020scalinglaws} of diffusion model RLFT through extensive experiments.

\subsection{Preliminaries on DDPO}
\paragraph{Markov Decision Process}
To use RL for finetuning, we need to formulate the task as a Markov Decision Process (MDP).
In a MDP, an agent interacts with the environment at discrete timesteps;
at each timestep $t$, the agent is at a state $s_t$, takes an action $a_t$ according to its policy $\pi(a_t|s_t)$, receives a reward $r_t$, and transitions to the next state $s_{t+1}$.
Following denoising diffusion policy optimization (DDPO)~\cite{black2023DDPO}, the denoising process of a diffusion model is formulated as a multi-step MDP:
\begin{align}
    s_t&=(c, t, x_t),\\
    a_t&=x_{t-1},\\
    \pi(a_t|s_t)&=p_\theta(x_{t-1}|c, t, x_t),\\
    r(s_t, a_t)&=
    \begin{cases}
    r(x_0,c) & \text{if } t=0,\\
    0 & \text{otherwise},
    \end{cases}\\
    r(x_0,c) &= -\text{MRC}(x_0)
\end{align}
where each denoising step is a timestep, $c$ is the context, i.e. the text prompt, $x_t$ is the image being denoised at step $t$, $p_\theta$ is the diffusion model being finetuned, $x_T$ is the initial noisy image, $x_0$ is the fully denoised image, and $r(x_0,c)$ is the negative \metricname (\appx\cref{lst:mrc}) computed on the fully denoised image.

\paragraph{Policy Gradient}
In order to optimize the model with respect to the reward function, a family of RL algorithms, known as policy gradient methods, are commonly adopted, such as REINFORCE~\cite{williams1992REINFORCE} and Proximal Policy Optimization (PPO)~\cite{schulman2017ppo}.
$\text{DDPO}_{\text{SF}}$ is based on the vanilla policy gradient algorithm, REINFORCE~\cite{williams1992REINFORCE}, also known as the Score Function (SF) of diffusion models.
On the other hand, $\text{DDPO}_{\text{IS}}$ builds upon PPO~\cite{schulman2017ppo} and conducts multiple optimization steps per round of data using an importance sampling (IS) estimator and importance weight clipping.

As a common practice to reduce the variance of the policy gradients~\cite{mnih2016a3c}, DDPO~\cite{black2023DDPO} uses the advantages (\cref{eq:A_r}), which are rewards normalized to have zero mean and unit variance, instead of directly using the rewards.
Specifically, the mean and standard deviation statistics of the rewards are tracked for each prompt $c$:
\begin{align}
    A_r(x_0,c) = \frac{r(x_0,c) - \mu_r(c)}{\sigma_r(c)}
    \label{eq:A_r}
\end{align}
DDPO's~\cite{black2023DDPO} reward-normalizing advantage replaces the value model that is more widely adopted in PPO-based~\cite{schulman2017ppo} RLHF methods~\cite{ouyang2022InstructGPT, yao2023deepspeedchat, vonwerra2022trl}. 
This is similar to~\cite{li2023remax}, which shows that the value model creates unnecessary computation cost that can be replaced with a simpler advantage formulation. 

By using the advantage term $A_r$ (\cref{eq:A_r}) in place of the reward $r$, the $\text{DDPO}_{\text{SF}}$ policy gradient function is:
\begin{align}
    \hat{g}_{\text{SF}} = \mathbb{E}\left[\sum_{t=0}^{T} \nabla_{\theta} \log p_{\theta}(x_{t-1} | c, t, x_t) A_r(x_0,c)\right]
    \label{eq:DDPO_SF}
\end{align}
where the expectation is taken over data generated by the policy $\pi_{\theta}$ with the parameters $\theta$.
The log probability $\log p_\theta(x_{t-1} | c, t, x_t)$ can be easily obtained since the policy is an isotropic Gaussian distribution when using the DDIM sampler~\cite{black2023DDPO,song2022ddim}.
The $\text{DDPO}_{\text{IS}}$ function (\appx\cref{eq:DDPO_IS}) has an additional importance sampling term than~\cref{eq:DDPO_SF}.

Black \etal~\cite{black2023DDPO} choose $\text{DDPO}_{\text{IS}}$ as the default policy gradient function, because it exhibits better sample efficiency than $\text{DDPO}_{\text{SF}}$ (Fig.~4 of~\cite{black2023DDPO}).
Such choice is consistent with the use of PPO~\cite{schulman2017ppo} in Large Language Model (LLM) Reinforcement Learning from Human Feedback (RLHF) literature~\cite{ouyang2022InstructGPT, yao2023deepspeedchat,vonwerra2022trl,bai2022helpfulharmlessRLHF,bai2022RLAIF}.

\begin{figure}
\centering
    \includegraphics[width=0.4\textwidth]{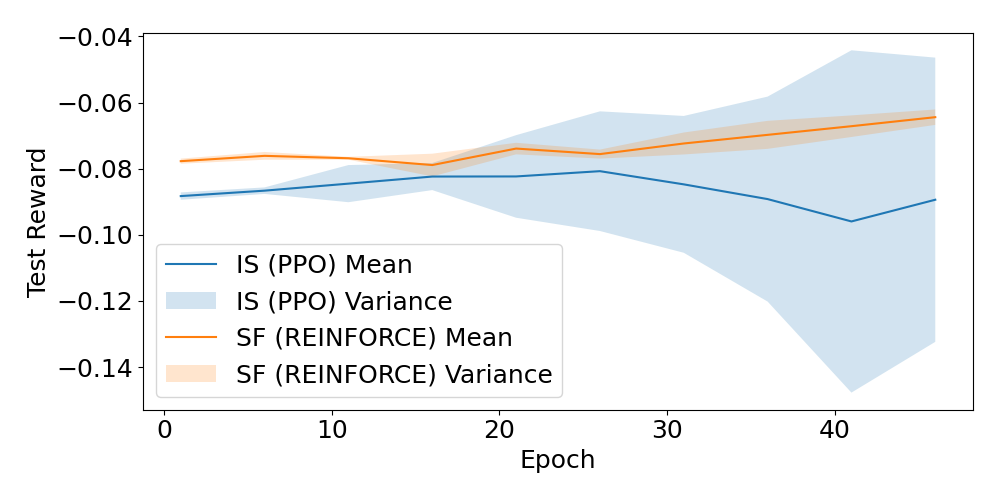}
    \vspace{-3mm}
    \caption{
    Comparing the IS and the SF versions of Carve3D reward curves on the testing set over 4 different random seeds.
    The IS version produces reward curves with high variance, including two runs that fails.
    In contrast, all runs of the SF version stably produces reward curves with low variance.
    \vspace{-3mm}
    }
    \label{fig:IS_vs_SF_variance}
\end{figure}

\begin{figure}
    \centering
    \includegraphics[width=0.4\textwidth]{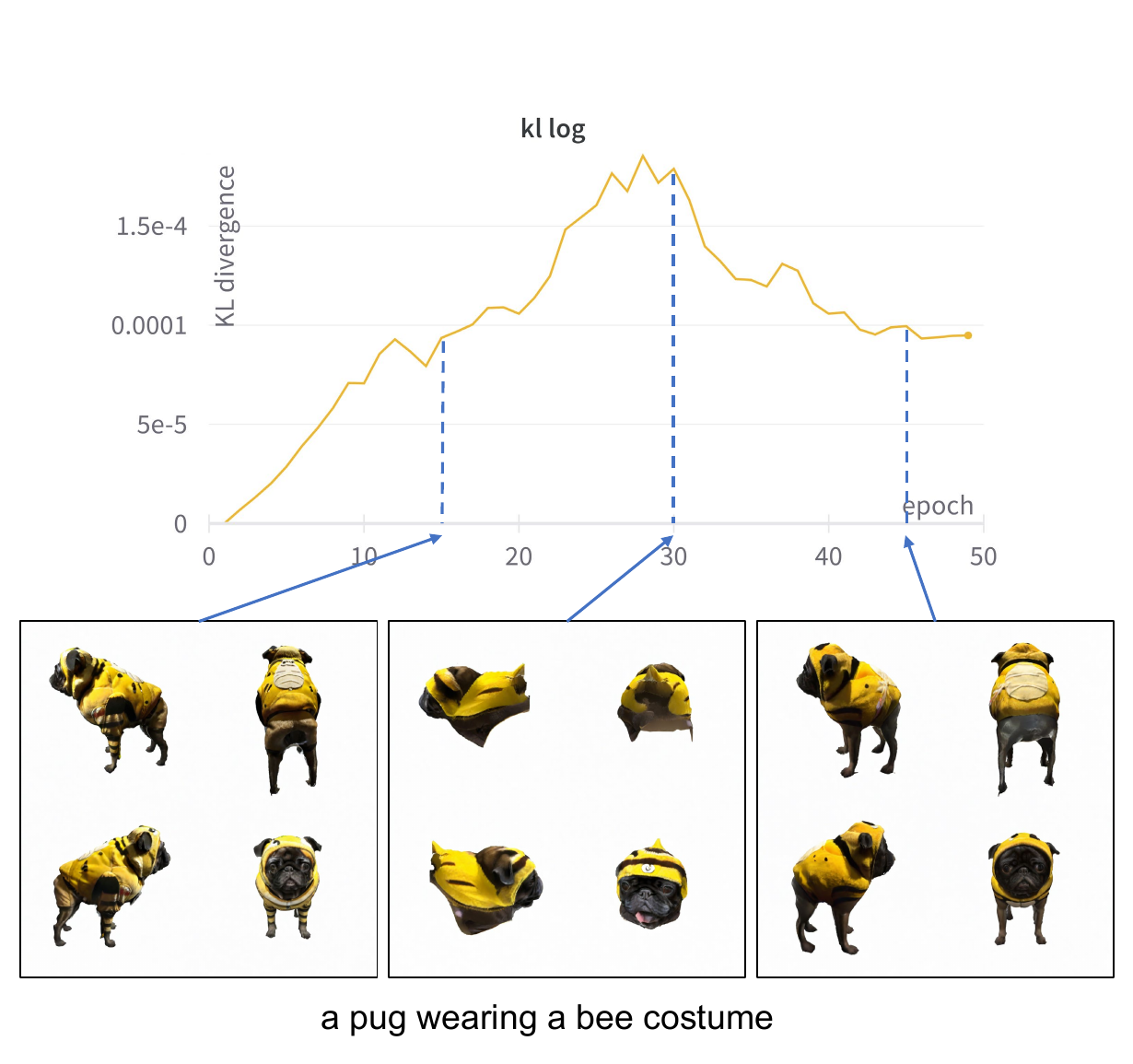}
    \vspace{-3mm}
    \caption{
    We observe qualitative correlation between the KL value and the prompt alignment degradation.
    Despite being distant in the finetuning process, epoch 15 and epoch 45, which have lower KL divergence to the base model, generates prompts that align better with the prompts.
    On the other hand, epoch 30, which has much higher KL divergence from the base model, generates results with broken identity, i.e. the body of the pug is missing.
    \vspace{-3mm}
    }
    \label{fig:kl_degradation}
\end{figure}

\subsection{Improvements over DDPO}
\label{par:RL:improvements}

While RLFT with the default $\text{DDPO}_{\text{IS}}$ and our \metricname can enhance the 3D consistency of multi-view diffusion models, it still faces challenges regarding training stability, the shift of output distributions, and an unclear training scale setting to achieve optimal rewards with minimal distribution shift.
To address these issues, we propose three improvements over DDPO~\cite{black2023DDPO} in this section.
Given the universal nature of these challenges in RLFT, our enhancements may offer broader applicability across various tasks.

\subsubsection{Pure On-policy Training}
Training stability is a major challenge in both RLFT~\cite{casper2023RLHFproblems, zheng2023RLHFsecrets} and traditional RL~\cite{eimer2023rlhyperparameters}.
With the default $\text{DDPO}_{\text{IS}}$, our training process is evidently unstable, as shown in~\cref{fig:IS_vs_SF_variance}.
Training experiments with different random seeds or a slight change of hyperparameters can lead to different reward curves and qualitative results.
This complicates the training result evaluation as we cannot distinguish meaningful improvement or deterioration from the variance introduced by random seed.

We argue that such high variance is derived from the multi-step update in $\text{DDPO}_{\text{IS}}$~\cite{black2023DDPO}, originally proposed in PPO~\cite{schulman2017ppo}.
While it theoretically allows for better sample efficiency similar to off-policy methods~\cite{schulman2017ppo}, it also causes the training to be more unstable and the reward curves to be more variant, because it uses data collected with the older policy to update the newer policy.
Due to the undesirable consequences of training instability, we adopt the pure on-policy variant $\text{DDPO}_{\text{SF}}$, discarding the multi-step update from PPO (\appx\cref{eq:DDPO_IS}).
As shown in~\cref{fig:IS_vs_SF_variance}, $\text{DDPO}_{\text{SF}}$ significantly improves the training stability of our RLFT, while maintaining a comparable sample efficiency as the default $\text{DDPO}_{\text{IS}}$.

Diverging from DDPO~\cite{black2023DDPO} and most LLM RLHF literature~\cite{ouyang2022InstructGPT, yao2023deepspeedchat,vonwerra2022trl,bai2022helpfulharmlessRLHF,bai2022RLAIF}, we choose REINFORCE~\cite{williams1992REINFORCE} ($\text{DDPO}_{\text{SF}}$) over PPO~\cite{schulman2017ppo} ($\text{DDPO}_{\text{IS}}$) for its superior training stability.
We provide two hypotheses behind our surprising finding in~\cref{sec:suppl:hypotheses}, including the difficulty of the task reward function and the size of the model being finetuned.
The favored use of REINFORCE~\cite{williams1992REINFORCE} over PPO~\cite{schulman2017ppo} could apply to broader scenarios that meet these two conditions. 
We leave the verification of our hypotheses as future work.

\subsubsection{KL Divergence Regularization}
\label{sec:RL:KL}
In RLFT methods, distribution shift (also known as reward overoptimization) can lead to low-quality results,
such as cartoon-like, less realistic style~\cite{black2023DDPO} or oversaturated colors and unnatural shape~\cite{fan2023dpok}, despite achieving high rewards.
In our case, we observe this as degradation of diversity, texture details and prompt alignment after prolonged RLFT with the \metricname reward. 
Previous methods~\cite{ouyang2022InstructGPT, fan2023dpok} suggest mitigating reward overoptimization by incorporating a penalty on the Kullback–Leibler (KL) divergence between the log probabilities of the outputs from the base and the finetuned models.
In our case, the base model is Instant3D-10K~\cite{li2023instant3d} without any additional finetuning.
By plotting the KL divergence values during finetuning, we also find that KL divergence correlates to the reward overoptimization problem (\cref{fig:kl_degradation}), suggesting us to adopt KL divergence regularization.

Following the widely adopted implementation in LLM RLHF~\cite{yao2023deepspeedchat, ouyang2022InstructGPT}, we incorporate KL penalty into the reward function. %
Subtraction of the log probabilities is commonly used to approximate the full KL divergence~\cite{vonwerra2022trl, yao2023deepspeedchat}:
\begin{align}
    \text{KL}\left( \log{p_\theta(x_{0}|c,T,x_T)} || \log{p_{\theta_\text{base}}(x_{0}|c,T,x_T)}\right) \nonumber \\
    = \sum_{t=0}^T{\frac{\log{p_\theta(x_{t-1}|c,t,x_t)} - \log{p_{\theta_\text{base}}(x_{t-1}|c,t,x_t)}}{T+1}}
    \label{eq:kl_computation}
\end{align}
where $p_{\theta_\text{base}}$ is the base model.
We will denote this approximated KL divergence term as $\text{KL}(x_0|c,x_T)$ for clarity in presentation. 

KL divergence values starts at $0$ and unavoidably increases as finetuning proceeds, making it hard to determine an optimal coefficient for the penalty term.
To enable a steady KL divergence regularization throughout the finetuning process, we propose to normalize the KL divergence penalty term.
This normalization ensures that the gradient consistently favors low-KL-divergence, high-reward samples, even in the early stages when KL divergence is still low compared to the later stages.
We extend DDPO's~\cite{black2023DDPO} per prompt stat tracking to also track the mean and standard deviation statistics of the KL divergence term in order to to normalize it:
\begin{align}
    A_\text{KL}(x_0,c) = \frac{\text{KL}(x_0|c,x_T) - \mu_\text{KL}(c)}{\sigma_\text{KL}(c)}.
    \label{eq:kl_normalized}
\end{align}
Our advantage terms now consist of both the normalized reward (\cref{eq:A_r}) and the normalized KL divergence (\cref{eq:kl_normalized}).
Our final policy gradient function, used in our experiments, is a combination of~\cref{eq:DDPO_SF,eq:kl_normalized}
\begin{align}
    \hat{g}_{\text{SF,KL}} = \mathbb{E}\left[\sum_{t=0}^{T} \nabla_{\theta} \log p_{\theta}(x_{t-1} | c, t, x_t) \right. \nonumber \\
    \left. \vphantom{\sum_{t=0}^{T}} \cdot \left(\alpha A_r(x_0,c) - \beta A_\text{KL}(x_0,c)\right)\right]
    \label{eq:DDPO_SF_KL}
\end{align}
where $\alpha$ and $\beta$ are the coefficients for the reward advantage and the KL advantage, respectively.

\begin{figure}
    \centering
    \includegraphics[width=0.47\textwidth]{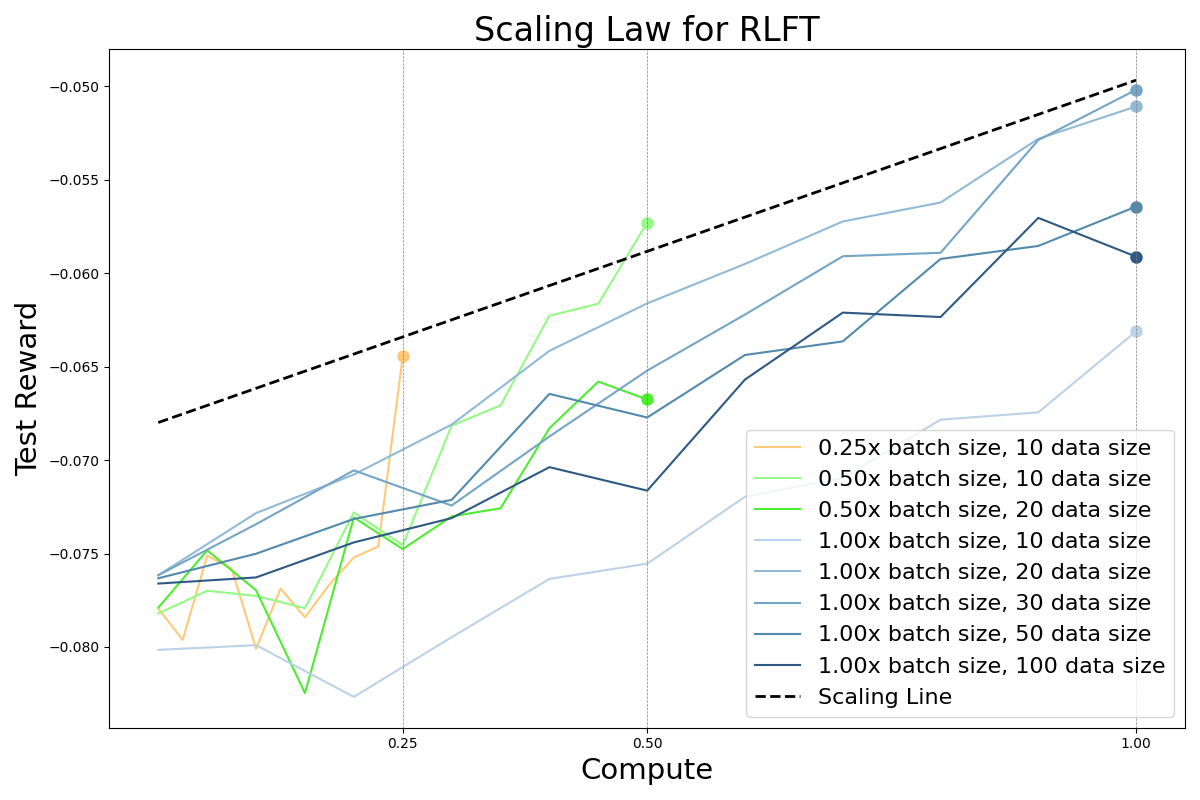}
    \vspace{-2mm}
    \caption{
    Scaling law for Carve3D's diffusion model RLFT algorithm.
    When we scale up the amount of compute, the model improves its reward smoothly under the optimal data size.
    The amount of compute scales linearly with respect to the batch size.
    The reward curves also become more stable (less variant) with a larger batch size.
    The reward curves are reported up to epoch $50$.
    }
    \label{fig:data_vs_batch}
    \vspace{-2mm}
\end{figure}
\subsubsection{Scaling Laws for RLFT}
\label{sec:RL:scale}

The training of Reinforcement Learning (RL) is highly sensitive to the chosen scale setting~\cite{eimer2023rlhyperparameters}, impacting various results, including the final converged reward.
Through the scaling laws identified from extensive experiments, we scale up the amount of compute (equivalent to scaling up the batch size in our case) to achieve the optimal reward.
Although our scaling experiments are only conducted with the multi-view consistency task, our insights into the scaling laws of diffusion model RLFT are general and can be beneficial in broader scenarios.

\paragraph{Compute and Batch Size}
The reward curves from our experiments demonstrate a positive scaling law of the model's reward at epoch 50 with respect to the amount of compute (\cref{fig:data_vs_batch}); 
the scaled up compute brings smooth improvement to the model's reward, under the optimal data sizes at each batch size.
Note that the amount of compute scales directly with respect to the batch size.

\paragraph{Data Size}
The model's reward does not directly scale with the data size but there exists a more complicated relationship between them. 
As shown in~\cref{fig:data_vs_batch}, the optimal data size at each batch size grows as the batch size get larger, indicating that both factors need to be scaled up in tandem;
after the optimal data size, naively continuing to scale up the data size actually reduces the model's reward.
Surprisingly, even when trained on a prompt set as small as a size of $10$, the model still shows generalization to the testing prompts.
We choose data size of 30 with batch size 768 in our final experiments as it achieves the highest reward in our analysis.

\paragraph{Training Epochs}
With the pure on-policy $\text{DDPO}_{\text{SF}}$ (REINFORCE~\cite{williams1992REINFORCE}), the model steadily and smoothly improves its rewards throughout the finetuning process, meaning that more training epochs constantly lead to higher reward.
However, from our qualitative results, we also observe worse distribution shift, e.g. the degradation of prompt alignment and texture details, as training epoch increases. 
Due to the correlation between KL divergence and the quality degradation (\cref{fig:kl_degradation}), we stop the finetuning early when a predefined KL divergence threshold is reached.
This threshold is empirically chosen based on qualitative results.
For fair comparisons, we report the reward curves up to epoch 50 in~\cref{fig:data_vs_batch}.
See~\cref{sec:suppl:epoch} for the definition of epoch in RLFT, which is different from that in other contexts, and~\cref{sec:suppl:training_data} for our training data generation and curation.

\begin{figure}
    \centering
    \includegraphics[width=0.45\textwidth]{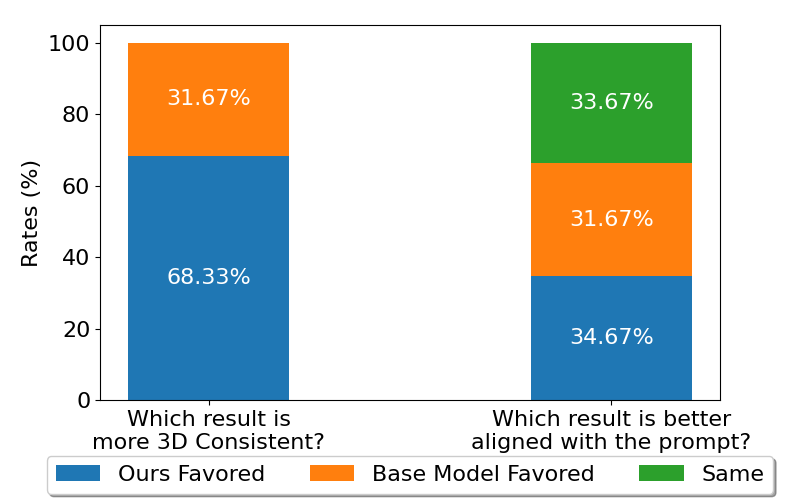}
    \caption{
    We conducted a user study with 20 randomly selected testing prompts and the corresponding outputs from both the base and fine-tuned model.
    $15$ participants took part in the study, with a majority favoring the 3D consistency of our fine-tuned model.
    Opinions are evenly split on which has better prompt alignment.}
    \vspace{-3mm}
    \label{fig:results:user_study}
\end{figure}
\section{Experiments}

In this section, we evaluate the Carve3D Model (Carve3DM), obtained by applying the Carve3D Reinforcement Learning Finetuning (RLFT) algorithm on Instant3D~\cite{li2023instant3d}.
See~\cref{sec:suppl:training_details} for our training details,~\cref{sec:suppl:experiment_details} for our experiment settings and~\cref{sec:suppl:ablation} for our ablation studies.

\subsection{Comparison with Base Model and Longer SFT}
\label{sec:results:compare}
As shown in~\cref{fig:IS_vs_SF_variance}, Carve3DM's Multi-view Reconstruction Consistency (MRC) reward steadily improves on the training set.
We aim to further answer the following questions:

\begin{enumerate}[leftmargin=0.8cm]
    \item Does Carve3DM's improved MRC generalize to the testing set?
    \item Quantitatively, is Carve3DM more consistent than the base model and the model with more Supervised Finetuning (SFT) steps?
    \item Qualitatively, does Carve3DM sacrifice the diversity, prompt alignment, and texture details of the base model?
\end{enumerate}

We quantitatively and qualitatively compare Carve3DM with Instant3D models with 10K, 20K, and 100K SFT steps, where Instant3D-10K is the default model in~\cite{li2023instant3d}.

\begin{table}[h]
\small
\centering
\begin{tabular}{|l|c|c|}
\hline
\textbf{}               & \textbf{Avg \metricname on Testing Set} $\downarrow$ \\ \hline
MVDream     & $0.1222$        \\ \hline
Instant3D-10K (Base)     & $0.0892$        \\ \hline
Instant3D-20K     & $0.0795$        \\ \hline
Instant3D-100K     & $0.0685$        \\ \hline
\textbf{Carve3DM (Ours)} & \textbf{0.0606}  \\ \hline
Zero123++ & 0.0700  \\ \hline
SyncDreamer & 0.1018  \\ \hline
\end{tabular}
\caption{
    Quantitative comparison of MVDream~\cite{shi2023MVDream}, Instant3D~\cite{li2023instant3d} with 10K (the base model), 20K, and 100K SFT steps, Carve3DM (ours, finetuned from Instant3D-10K), Zero123++~\cite{shi2023zero123plus}, and SyncDreamer~\cite{liu2023syncdreamer}.
    We evaluate them by generating 4 outputs for each of the 414 text prompts in the DreamFusion~\cite{poole2022dreamfusion} testing set.
    We let Zero123++ and SyncDreamer to use one of Carve3DM's output multi-view images as their input image conditioning.
    Carve3DM achieves substantially better MRC than all baselines, indicating its superior multi-view consistency.
}
\label{tab:suppl:testing_mrc}
\vspace{-2mm}
\end{table}

\paragraph{Quantitative Comparison and Generalization}
As shown in~\cref{tab:suppl:testing_mrc}, 
when evaluated on the testing set, \methodname achieves substantially improved \metricname over the base model. 
More SFT steps indeed provides better multi-view consistency and achieves better MRC, i.e. Instant3D's 100K version is better than the 20K version, which is better than the 10K version. 
However, Carve3DM still outperforms even the most consistent 100K version of Instant3D by a notable gap.
This suggests that the explicit multi-view consistency objective in MRC, paired with our RLFT algorithm, can improve the model's consistency more efficiently than SFT.

Furthermore, although our RLFT only uses 30 training prompts, it brings multi-view consistency improvement that generalizes to the testing set containing 415 prompts.
Such generalization, also observed in~\cite{black2023DDPO,ouyang2022InstructGPT}, is likely derived from the strong knowledge from the base model.

\paragraph{Multi-view Consistency and NeRF Artifacts}
\appx\cref{fig:suppl:qualitative} shows the improved multi-view consistency and the resulting Neural Radiance Field (NeRF) reconstruction quality.
While the multi-view images generated by the base model may be inconsistent, causing artifacts such as floaters and broken geometry, Carve3D can fix such inconsistency in the multi-view images and produce NeRF with clean geometry, free of artifacts.
In \appx\cref{fig:suppl:rl_vs_sft}, Carve3DM continues to show superior multi-view consistency and reduced NeRF artifacts, but such improvement is less obvious when compared to the 20K and 100K version of Instant3D~\cite{li2023instant3d}, similar to our quantitative results in~\cref{tab:suppl:testing_mrc}.

\paragraph{Prompt Alignment and Texture Details}
By virtue of our RLFT with KL-divergence regularization (\cref{sec:RL:KL}), which prevents distribution shift, and our curation of training prompt dataset (\cref{sec:suppl:training_data}),
Carve3DM preserves the prompt alignment and the texture details of the base model, as shown in~\appx\cref{fig:suppl:qualitative}.
On the other hand, longer SFT causes additional distribution shift in Instant3D~\cite{li2023instant3d} from the base SDXL~\cite{podell2023sdxl} towards the SFT training set~\cite{deitke2022objaverse}.
As shown in~\appx\cref{fig:suppl:rl_vs_sft}, Instant3D-20K and Instant3D-100K exhibits degradation in diversity, realism, and level of detailed textures.
This quality degradation with longer SFT is also observed in~\cite{li2023instant3d}.

\paragraph{Diversity}
As shown in~\appx\cref{fig:suppl:diversity}, Carve3DM preserves the generation diversity of the base model.
This owes to our RLFT process, which repeatedly samples different initial noises for the diffusion model to generate diverse results (\cref{fig:overview}).

\subsection{Comparison with Existing Methods}
\label{sec:suppl:mrc_mvdream}
We further compare Carve3DM with the text-conditioned multi-view diffusion model, MVDream~\cite{shi2023MVDream}, and two image-conditioned models, Zero123++~\cite{shi2023zero123plus} and SyncDreamer~\cite{liu2023syncdreamer}.
As shown in~\cref{tab:suppl:testing_mrc} and~\appx\cref{fig:suppl:rl_vs_sft}, Carve3DM's outputs have notably better multi-view consistency, realism, and level of details than the three baselines.

\subsection{User Study}
In addition to the quantitative and qualitative comparisons in~\cref{sec:results:compare}, we conduct a user study to further understand the qualitative results of Carve3DM when perceived by human.
To run the user study, we \textit{randomly} selected 20 unseen testing prompts.
For each text prompt, we generated a pair of data from both the base and the finetuned models with the same initial noise. 
Then, we provided both the tiled 4-view image and the turntable video of the reconstructed NeRF to participants and asked them the following two questions:
(1) Which result is more 3D-consistent? and 
(2) Which result is better aligned with the prompt?
As shown in~\cref{fig:results:user_study}, 68.33\% of participants believe that Carve3DM's generated results are more 3D consistent than that of the base model~\cite{li2023instant3d}.
Given that the multi-view consistency in the base model has already been much improved with SFT~\footnote{Please see \href{https://jiahao.ai/instant3d/}{https://jiahao.ai/instant3d/} for base model's 3D consistency},
the nearly $37\%$ gain in human preference introduced by \methodname on \textit{randomly} selected testing prompts is impressive.
Furthermore, participants find that Carve3DM exhibits similar prompt alignment as Instant3D.
The preservation of alignment can be attributed to the KL divergence regularization (\cref{sec:RL:KL}) and early stopping the RLFT regarding KL divergence (\cref{sec:RL:scale}).

\section{Conclusion}

In this paper, we propose \methodname, a Reinforcement Learning Finetuning algorithm to improve the reconstruction consistency of 2D multi-view diffusion models.
\methodname relies on \metricname, a novel metric that measures the reconstruction consistency by comparing multi-view images with their corresponding NeRF renderings at the same viewpoints.
The resulting model, Carve3DM, demonstrates substantially improved multi-view consistency and NeRF quality without sacrificing the prompt alignment, texture details, or prompt alignment of the base model.
Our results conclude that pairing SFT with Carve3D's RLFT is essential for developing multi-view-consistent diffusion models.

\ignore{
Broader Impacts: promote RL finetuning, alignment research, could be used to make safer, better aligned, reduced bias, harm diffusion models.

Metric could be used to evaluate/compare existing text-to-multi-view Diffusion Models. 
It could also be used to monitor the fineutning process and picking the best checkpoint.

Limitation: loss of high-frequency detail, due to limitation of Sparse-view LRM.
}
{
    \small
    \bibliographystyle{ieeenat_fullname}
    \bibliography{main}
}

\appendix
\setcounter{page}{1}

\begin{listing*}
\begin{minted}[breaklines, fontsize=\small]{python}
def compute_mrc(ori_views, ori_cam_poses, lrm, lpips, resize_res):
    nerf = lrm(ori_views, ori_cam_poses)
    nerf_views = nerf.render(ori_cam_poses)
    square_bbox = compute_square_bbox(ori_views) # bounding box coordinates for each view
    x_min, y_min, x_max, y_max = square_bbox
    ori_views_bbox = [resize(o[:, y_min:y_max + 1, x_min:x_max + 1], resize_res) for o in ori_views]
    nerf_views_bbox = [resize(n[:, y_min:y_max + 1, x_min:x_max + 1], resize_res) for n in nerf_views]
    mrc = lpips(ori_views_bbox, nerf_views_bbox).mean()
    return mrc
\end{minted}
\caption{
Pseudo code for our MRC implementation.
ori\_views and ori\_cam\_poses are the multi-view images to be evaluated and their camera poses.
lrm is the sparse-view LRM~\cite{li2023instant3d,hong2023lrm}.
lpips the the LPIPS~\cite{zhang2018lpips} metric.
resize\_res is a fixed resolution to which we resize the bounding box patches.
}
\label{lst:mrc}
\end{listing*}
\section{Appendix Summary}
In the appendix, we cover additional details of DDPO~\cite{black2023DDPO} and Reinforcement Learning Finetuning (RLFT) (\cref{sec:suppl:ddpo}), training data details (\cref{sec:suppl:training_data}), training details (\cref{sec:suppl:training_details}), experiment details (\cref{sec:suppl:experiment_details}), additional Multi-view Reconstruction Consistency (MRC) metric experiments (\cref{sec:suppl:mrc_exp}), ablation studies (\cref{sec:suppl:ablation}), and broader impacts and future work (\cref{sec:suppl:limitation}).

\section{Additional Details of DDPO and RLFT}
\label{sec:suppl:ddpo}

\subsection{Definitions}
\label{sec:suppl:epoch}
Following~\cite{schulman2017ppo,ouyang2022InstructGPT, yao2023deepspeedchat, black2023DDPO}, an \textit{epoch} is defined as one round of data collection (\textit{sampling}), which may consists multiple PPO~\cite{schulman2017ppo} update steps (\textit{training}), as discussed in~\cref{eq:DDPO_IS,par:RL:improvements}.
This definition of ``epoch'' is different from the meaning in supervised learning which usually refers to go through all data once. 
Since we opt for using pure on-policy training (vanilla policy gradient), as discussed in~\cref{par:RL:improvements}, we only do one training step per sampling step, and thus our sampling batch size and training batch size are equal.

\subsection{$\text{DDPO}_{\text{IS}}$ Policy Gradient Function}
\label{sec:suppl:DDPO_IS}
By using the advantage term (\cref{eq:A_r}) in place of the reward, the $\text{DDPO}_{\text{IS}}$ policy gradient function is:
\begin{align}
    \hat{g}_{\text{IS}} = \mathbb{E}&\left[\sum_{t=0}^{T} \frac{p_\theta(x_{t-1} | c, t, x_t)}{p_{\theta_\text{old}}(x_{t-1} | c, t, x_t)} \right. \nonumber\\
    &\left. \vphantom{\sum_{t=0}^{T}} \cdot \nabla_{\theta} \log p_\theta(x_{t-1} | c, t, x_t) A_r(x_0,c)\right]
    \label{eq:DDPO_IS}
\end{align}
where the expectation is taken over data generated by the policy $\pi_{\theta_\text{old}}$ with the parameters $\theta_\text{old}$.

\subsection{Hypotheses on Stability and Sample Efficiency}
\label{sec:suppl:hypotheses}
Diverging from DDPO~\cite{black2023DDPO} and most Large Language Model (LLM) Reinforcement Learning from Human Feedback (RLHF) literature~\cite{ouyang2022InstructGPT, yao2023deepspeedchat,vonwerra2022trl,bai2022helpfulharmlessRLHF,bai2022RLAIF}, we choose REINFORCE~\cite{williams1992REINFORCE} ($\text{DDPO}_{\text{SF}}$) over PPO~\cite{schulman2017ppo} ($\text{DDPO}_{\text{IS}}$) for its superior training stability.
We provide two hypotheses behind our surprising finding.

(1) Training stability is more vital than sample efficiency when the task reward function is more challenging.
When a reward function is more variant with respect to the model's output, it becomes more difficult for the model to discover the pattern of high-reward outputs and to improve its rewards.
The high-variance prompt alignment reward curves in Fig.~5 of DDPO~\cite{black2023DDPO} indicates the challenging nature of the prompt alignment task as opposed to the smooth reward curves for the aesthetics and compressibility tasks in Fig.~4 of DDPO~\cite{black2023DDPO}.

(2) The RL Finetuning (RLFT) sample efficiency is less important for a large model which requires less finetuning steps, as
demonstrated in studies of LLM instruction finetuning~\cite{chung2022scaling}.
Similarly, our RLFT on a 2.6B-parameter UNet from SDXL~\cite{podell2023sdxl} only takes 55 epochs, as opposed to DDPO's~\cite{black2023DDPO} RLFT on a 860M-parameter UNet from SD 1.4~\cite{rombach2022sd} using 200 epochs.
Therefore, the potential sample efficiency gain provided by the multi-step update of PPO~\cite{schulman2017ppo} gets outweighted by the training stability provided by REINFORCE~\cite{williams1992REINFORCE}.

The favorableness of REINFORCE~\cite{williams1992REINFORCE} could apply to broader scenarios that fits these two conditions. 
We leave the verification of our hypotheses as future work.

\section{Implementation Details}

\subsection{Training Data}
\label{sec:suppl:training_data}
An advantage of Reinforcement Learning Finetuning (RLFT) over Supervised Finetuning (SFT) is that, we can manually create a high-quality text prompts training set, while creating a dataset of diverse ground truth multi-view images for these high-quality text prompts is prohibitively expensive for SFT.
By relying on samples generated by the model itself to compute the reward and the loss, RLFT can optimize a model beyond the limitation of a dataset and preserves the diversity and the style of the base model.
In Carve3D, our training prompts preparation process involves two strategies.

\paragraph{Training Data Curation}
Instead of randomly sampling prompts from a dataset, we employ a data curation strategy where prompts with lowest rewards are selected.
Specifically, we run inference of the base model on a prompt dataset, generating four results per prompt, compute the MRC rewards for each result, and sort the prompts according to their average reward.
This is derived from observation that, for certain prompts, the model generates nearly optimal outputs with rewards close to the rewards of ground truth views of a 3D asset~\cite{deitke2022objaverse} (\cref{fig:method:consistency-plot}).
Thus, the curated lowest-reward prompts have substantial room for 3D consistency improvement and prevent learning stagnation.
This approach not only brings more efficient training but also provides a more generalized improvement in 3D consistency to the testing set.

\paragraph{Creating New Training Prompt Set}
The prompt dataset from DreamFusion~\cite{poole2022dreamfusion}, which contains 414 prompts and is commonly used as testing set.
To employ the DreamFusion prompt set also as our testing set, we create a new prompt dataset with ChatGPT4~\cite{openai_chatgpt}.
Following our training data curation strategy, we first sort the DreamFusion~\cite{poole2022dreamfusion} prompts according to their rewards attained by the base Instant3D~\cite{li2023instant3d} model.
We provide the sorted prompt set to ChatGPT4, and ask it to summarize the characteristics of the low-reward prompts by looking at the low-reward, median-reward, and high-reward prompts.
ChatGPT4 summarizes low-reward prompts to possess properties of ``complex and creative''.
We then ask it to generate 100 low-reward prompts that are both complex and creative, and another 100 low-reward prompts that are ``complex but not too creative".
For each set, again, we sort the prompts according to their rewards, and select those with the lowest rewards to be our training prompt set.
Our best results are obtained with the ``complex but not too creative" set.

\subsection{Training Details}
\label{sec:suppl:training_details}
All of our RL finetuning experiments are run on 6 AWS EC2 P4de nodes with 8 NVIDIA A100-SXM4-80GB GPUs, a total of 48 GPUs.
We use batch size of 768, which is 2x compared to that of DDPO.
One experiment takes $16.5$ hours to reach 55 epochs.
The number of finetuning epochs is determined by our KL-divergence early-stopping mechanism, which we empirically choose to be $3.2e-4$ according to the level of reward overoptimization shown on qualitative results.

We use minibatches of size 8 during sampling and 4 during training due to the limited GPU memory.
The total batch size of 768 is evenly distributed among each GPU, so that the per GPU batch size is 16.
The model samples two minibatches of size 8 on all GPUs to reach the total batch size.
Similarly, the model accumulates gradients over four minibatches of size 4 on all GPUs, before synchronizing the gradients and performing an optimizer step.
We use a per prompt stat tracker with windows of size 76, so that it roughly tracks all the rewards per prompt ever 3 epochs.
This is much larger than DDPO's default tracking window of size 32 for better training stability.
The coefficients for the advantage terms in~\cref{eq:DDPO_SF_KL} are $\alpha=1$ and $\beta=0.2$.

The rest of our RL finetuning setup follows DDPO~\cite{black2023DDPO,ddpo_pytorch}.
We use the AdamW~\cite{loshchilov2017adamw} optimizer with a fixed learning rate $3e-4$, $\beta_1=0.9$, $\beta_2=0.999$, $\epsilon=1e-8$ and a weight decay of $1e-4$. 
The high learning rate is paired with Low Rank Adaptation (LoRA)~\cite{hu2021lora} finetuning with rank 4, which significantly reduces the memory and computation requirements for finetuning.
We freeze all networks in the base model and set their precision to fp16, and only finetune the LoRA weights of the unet under fp32 via mixed precision training.

Our base text-to-multiview diffusion model setup follows Instant3D~\cite{li2023instant3d}, which uses the same architecure as SDXL~\cite{podell2023sdxl}.
It produces images of resolution 1024x1024, which contains four images of resolution 512x512 tiled in a 2-by-2 fashion.
Instant3D requires 100 denoising steps during inference, doubling the required computation than the default 50 steps for SDXL. 
It uses Classifier Free Guidance~\cite{ho2022cfg} with a scale of 5.0

Our code is mainly based on DDPO's~\cite{black2023DDPO} official implementation, the ddpo-pytorch~\cite{ddpo_pytorch} Github repository, which uses Hugginface diffusers~\cite{von-platen-etal-2022-diffusers} and PyTorch~\cite{paszke2019pytorch} libraries.
Our KL divergence regularization implementation is inspired by the codebases of DeepSpeedChat~\cite{yao2023deepspeedchat}, TRL~\cite{vonwerra2022trl}, and DPOK~\cite{fan2023dpok}.
We thank the authors of these repositories for releasing the high-quality implementations and promoting open-sourced research.
We are going to release the code for computing MRC and our improved DDPO implementation. 
However, due to the fact that Sparse View LRM and Instant3D do not plan to release their code, we have to leave these as empty, abstract functions in our released code.

\subsection{Experiment Details}
\label{sec:suppl:experiment_details}
All quantitative and qualitative experiments and the user study uses the results from evaluating each model on the DreamFusion~\cite{poole2022dreamfusion} testing prompt set, containing 415 prompts, for 4 times, which equals to 1660 results per model.

For fair qualitative comparisons, results from each model are generated from the the same initial noise.
Since Instant3D-20K and -100K and Carve3D are all finetuned from Instant3D-10K, their output tend to represent the same object when given the same initial noise (e.g.~\cref{fig:teaser,fig:kl_degradation,fig:suppl:rl_vs_sft}).

Since Zero123++~\cite{shi2023zero123plus} and SyncDreamer~\cite{liu2023syncdreamer} are image-to-multi-view diffusion models, we let them take one of Carve3D's output image and their input image-conditioning. 
Therefore, their output has the same level of prompt alignment, texture details, and diversity as Carve3D.

\section{Additional MRC Metric Experiments}
\label{sec:suppl:mrc_exp}

\paragraph{Distortion Types}
Here, we show the full results for the metric experiments for the inpainting distortion (\cref{fig:suppl:metricexp:inpaint}) discussed in~\cref{sec:metric:exp,fig:method:consistency,fig:method:consistency-plot}.
We also conduct metric experiments with other distortions types: azimuth rotation (\cref{fig:suppl:metricexp:azimuth}, and elevation rotation (\cref{fig:suppl:metricexp:elevation}).
In azimuth and elevation rotation, for one out of the four views, we rotate the object with an azimuth or elevation rotation by $3.6$ or $4$ degrees, before rendering that view, and also use the original camera extrinsic matrix as the input to Sparse View LRM~\cite{li2023instant3d,hong2023lrm}.
The quantitative results matches our expectations, where MRC, i.e. with LPIPS, monotonically decreases as we intentionally add more distortion.

\paragraph{LPIPS vs. Other Image Similarity Metrics}
Here, we compare substituting LPIPS~\cite{zhang2018lpips} with L1, L2, PSNR, and SSIM in the Multi-view Reconstruction Consistency (MRC) metric experiments on all distortion types.
In the inpainting distortion experiments (\cref{fig:suppl:metricexp:inpaint}), which is the most representative of diffusion model's inconsistencies, LPIPS is more linear than other pixel level image metrics.
In azimuth and elevation distortion experiments (\cref{fig:suppl:metricexp:azimuth,fig:suppl:metricexp:elevation}), all image metrics shows monotonically decreasing pattern, while pixel-level image metrics are more linear.
This is expected as the distortion is pixel-aligned and more structured.

\begin{figure}
    \centering
    \includegraphics[width=0.35\textwidth]{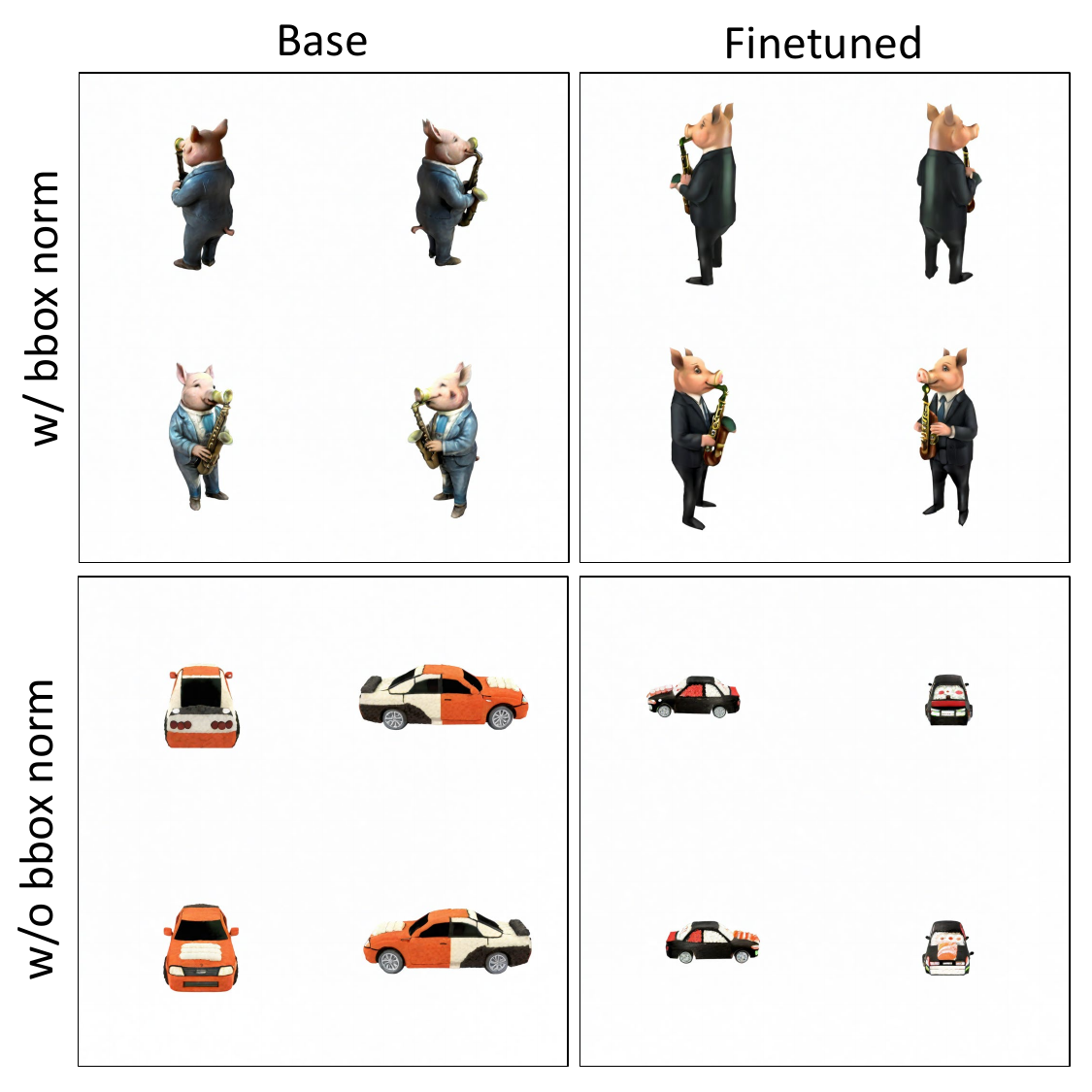}
    \caption{
    Alation study on the boundingbox normalization of LPIPS for \metricname evaluation.
    Top: with boundingbox normalization, the size of the foreground object is similar that of the base model. 
    Bottom: without boundingbox normalization, the size of the foreground object after RL finetuning is substantially smaller than that of the base model.
    }
    \label{fig:suppl:bbox_norm}
\end{figure}
\begin{figure}
\centering
\begin{subfigure}[b]{\textwidth}
   \includegraphics[width=0.4\textwidth]{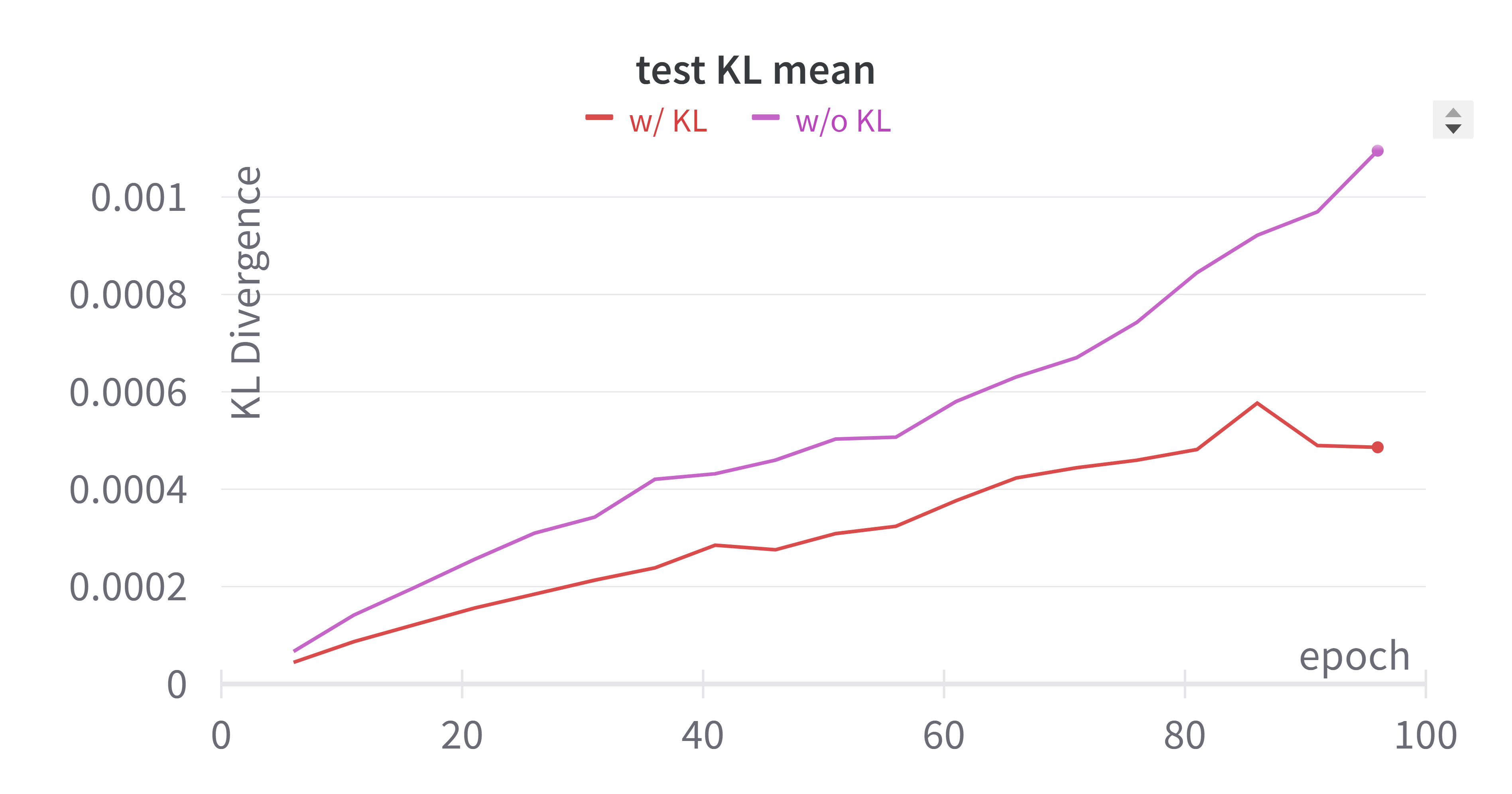}
\end{subfigure}

\begin{subfigure}[b]{\textwidth}
   \includegraphics[width=0.4\textwidth]{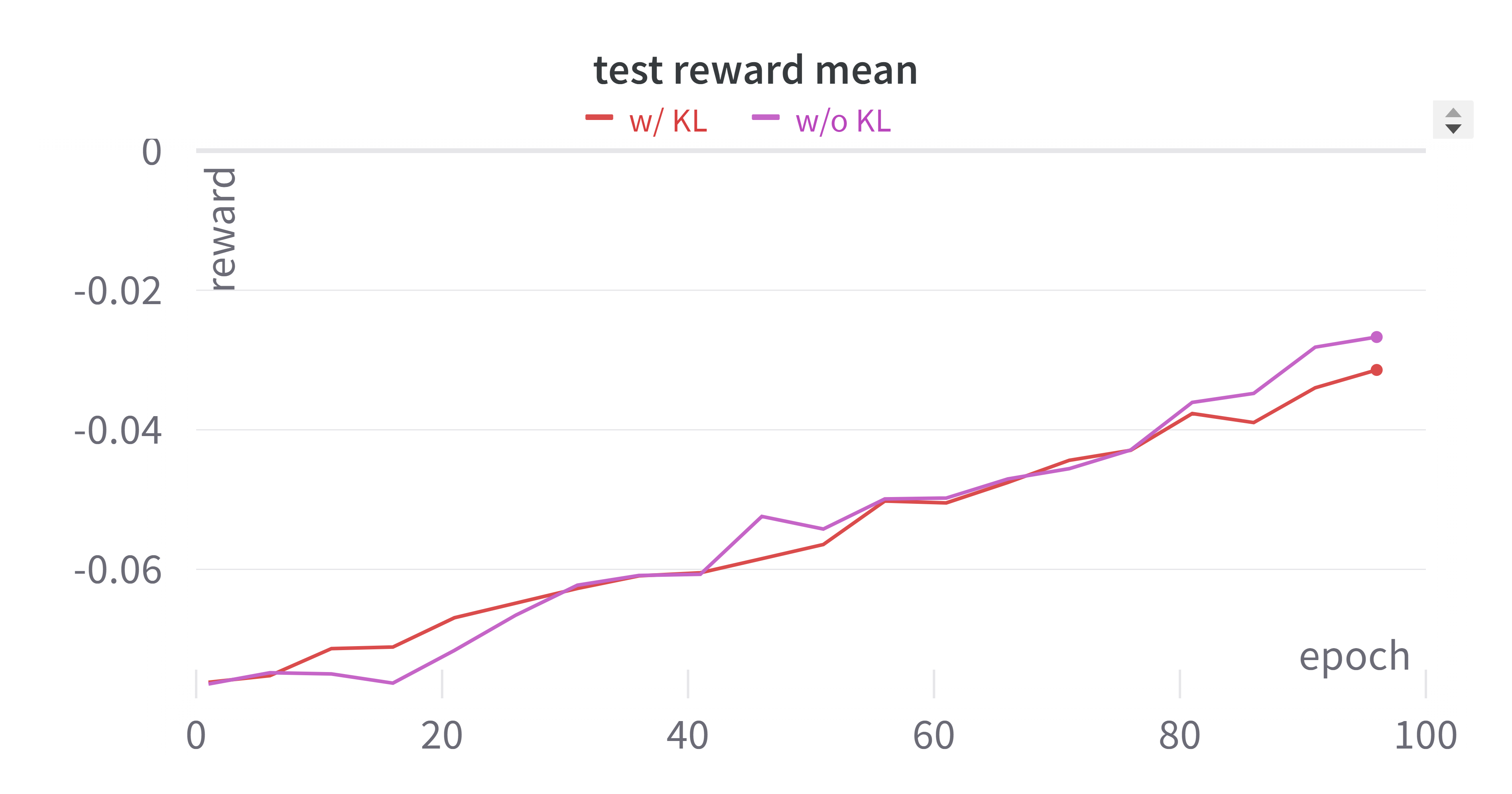}
\end{subfigure}
\caption{
    Ablation study on KL divergence regularization.
    Top: KL Divergence between the base model and the finetuned model on testing set. 
    Bottom: mean MRC reward on testing set.
    Our KL divergence regularization does not sacrifice the model's sample efficiency of the reward curve.
    The finetuned model has a lower KL divergence to the base model when using KL divergence regularization than without, which prevents degraded object identity and reduced texture details.
}
\label{fig:suppl:kl_ablation}
\end{figure}
\section{Ablation Studies}
\label{sec:suppl:ablation}
\paragraph{Bounding Box Normalization}
As shown in~\cref{fig:suppl:bbox_norm}, when the bounding box normalization is removed from MRC, the model would trivially increase the reward by reducing the size of the foreground object on the white background.
This would lead to the model generating images containing only the white background, after longer finetuning.
With bounding box normalization, the model would learn the harder task of improving the reconstruction consistency of the multi-view images.

\paragraph{KL Divergence Regularization}
As shown in~\cref{fig:suppl:kl_ablation}
Our KL divergence regularization does not sacrifice the model's efficiency on improving its reward.
Without KL divergence regularization, the KL divergence grows much faster. 
As discussed in~\cref{par:RL:improvements}, this leads to degraded object identity and loss of texture details.
\section{Broader Impacts and Future Work}
\label{sec:suppl:limitation}
Our Multi-view Reconstruction Consistency (\metricname) metric can serve as a valuable tool for evaluating any multi-view generative methods and guiding future developments in the field.
Although we only demonstrate our Reinforcement Learning Finetuning (RLFT) with MRC on one multi-view diffusion model~\cite{li2023instant3d}, it can be directly adapted to other text-to-multi-view diffusion models;
such adaptation only requires tuning a few hyperparameters related to the scaling laws for diffusion model RLFT (\cref{sec:RL:scale}).
Our surprising finding behind the choice of REINFORCE~\cite{williams1992REINFORCE} over PPO~\cite{schulman2017ppo} for better training stability could also be applied in broader RLFT scenarios.

As AI models grow more powerful, it becomes more important to evaluate and improve their safety and reduce their bias.
RLFT has been widely used for Large Language Model (LLM) alignment as it allows models to be finetuned with hard-to-specify objectives and its results are generalizable without undermining the base model's knowledge.
As the first work to use RLFT for text-to-3D and on diffusion models at the SDXL scale,
we hope Carve3D can inspire more alignment research in the computer vision community.

Carve3D is limited by the reconstruction quality of Sparse View Large Reconstruction Model (LRM)~\cite{li2023instant3d,hong2023lrm}.
Because its reconstruction is not perfect, this leads to non-zero MRC metric on GT views as shown in~\cref{fig:suppl:metricexp:inpaint,fig:suppl:metricexp:azimuth,fig:suppl:metricexp:elevation}. 
Due to this limitation of Sparse View LRM, Carve3D RL finetuned model can produce less high-frequency details than the base model in order to lower the image distance to the NeRF rendered views.
This might be solved by using a future sparse view reconstructor that can preserve more details or training a dedicated model for computing MRC.

Further increasing data size and batch size to further improve reconstruction consistency is possible.
However, in this work, we are limited by the high computation cost of SDXL~\cite{podell2023sdxl}, Instant3D's 100 denoising steps, and the high number of samples needed in DDPO. 
A few concurrent works could address this challenge.
It is possible to substantially reduce the computation cost by switching to Consistency Models for one/few-step inference (e.g., LCM-LoRA~\cite{luo2023lcmlora}).
In addition, we can also switch from DDPO to direct backpropagation of reward (e.g. Align-Prop~\cite{prabhudesai2023alignprop}, and DRaFT~\cite{clark2023DRaFT}) to reduce the number of samples needed. 
We leave these extensions as future work.
\begin{figure*}
    \centering
    \includegraphics[width=0.99\textwidth]{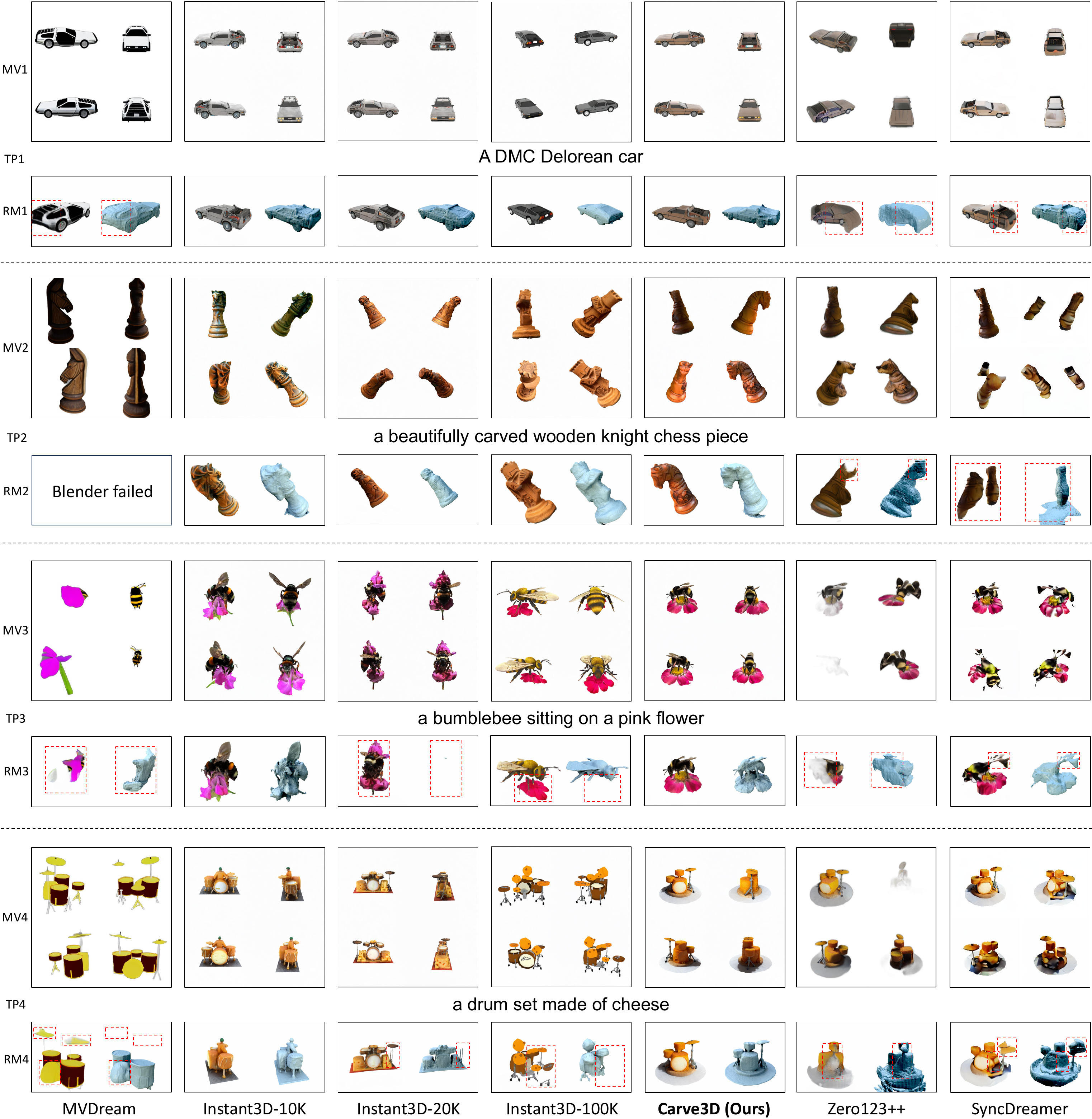}
    \caption{
        Qualitative comparison of MVDream~\cite{shi2023MVDream}, Instant3D~\cite{li2023instant3d} with 10K (the base model), 20K, and 100K SFT steps, Carve3D (ours, finetuned from Instant3D-10K), Zero123++~\cite{shi2023zero123plus}, and SyncDreamer~\cite{liu2023syncdreamer} (7 models in 7 columns) on 4 prompts (in 4 rows, numbered as 1-4, separated by dotted line).
        In each row, we show their generated multi-view images in the 2-by-2 grid (denoted as MV), reconstructed NeRF and extracted mesh (denoted as RM) when given the text prompt (denoted as TP).
        MVDream, Zero123++, and SyncDreamer generates inconsistent multi-view images and reconstruction artifacts (highlighted in red).
        For each result, we use the same randomly sampled initial noise for all models to ensure the comparison is fair.
        We let Zero123++ and SyncDreamer to use one of Carve3D's output multi-view images as their input image conditioning.
        Instant3D-10K, -20K, and -100K and Carve3D demonstrates progressively better multi-view consistency and reconstruction quality.
        Instant3D-10K, Carve3D, Zero123++, and SyncDreamer exhibits the best texture details and realism, whereas Instant3D-20K and -100K with prolonged SFT steps compromise those qualities.
    }
    \label{fig:suppl:rl_vs_sft}
\end{figure*}
\begin{figure*}
    \centering
    \includegraphics[width=0.99\textwidth]{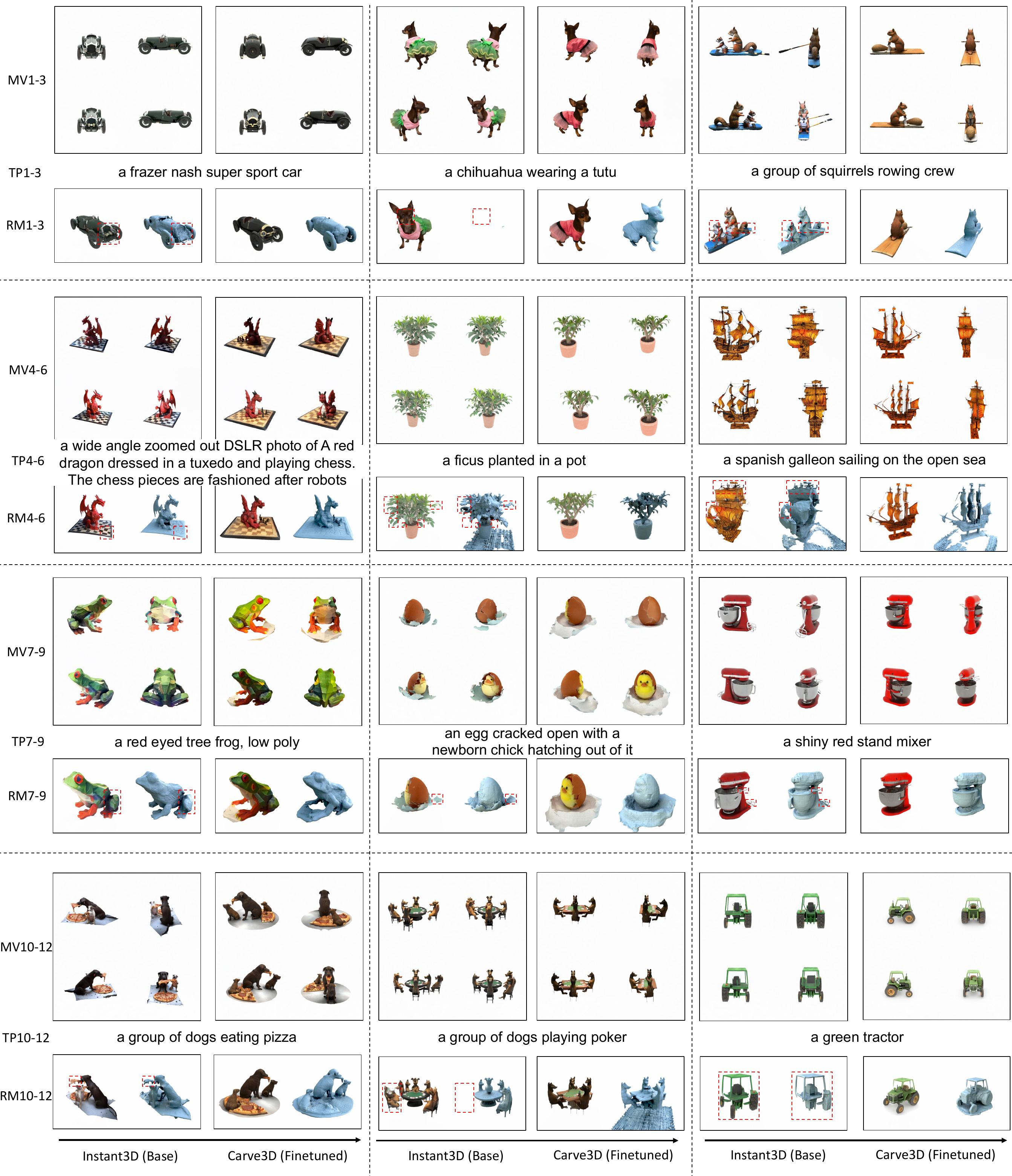}
    \caption{
    Qualitative comparison of Instant3D (the base model) and Carve3D (ours, finetuned from Instant3D) on $12$ prompts (in 12 blocks, numbered as 1-12, separated by dotted line).
    In each block, we show the their generated multi-view images in the 2-by-2 grid (denoted as MV), the reconstructed NeRF and the extracted mesh (denoted as RM) when given the text prompt (denoted as TP).
    For each result, we use the same randomly sampled initial noise for all models to ensure the comparison is fair.
    We draw red boxes on the NeRF and the extracted mesh to highlight the artifacts in the NeRF and the mesh, resulting from the inconsistencies in the multi-view images.
    Carve3D maintains the detailed texture and provides improved multi-view consistency and higher quality NeRF than the base Instant3D.
    }
    \label{fig:suppl:qualitative}
\end{figure*}
\begin{figure*}
    \centering
    \includegraphics[width=0.99\textwidth]{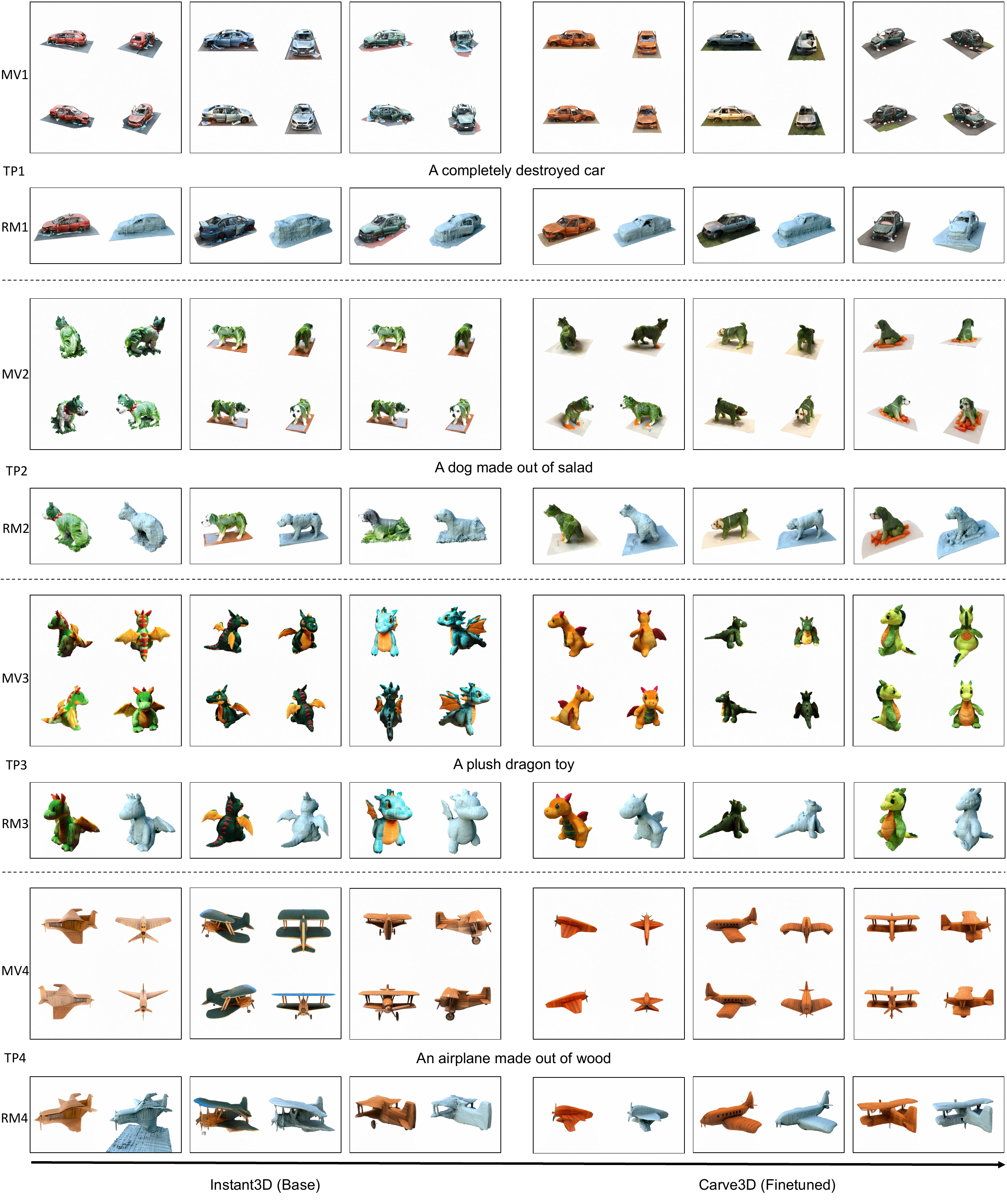}
    \caption{
        Qualitative comparison of Instant3D (the base model) and Carve3D (ours, finetuned from Instant3D) on 4 prompts (in 4 rows, numbered as 1-4, separated by the dotted line) demonstrating diversity. 
        In each row, we show 3 results from each model, including the generated multi-view images in the 2-by-2 grid (denoted as MV), the reconstructed NeRF and the extracted mesh (denoted as bottom) when given the prompt (denoted as middle).
        For each result, we use the same randomly sampled initial noise for all models to ensure the comparison is fair.
        Our RLFT maintains the diversity of the base Instant3D model, while improving the consistency. 
    }
    \label{fig:suppl:diversity}
\end{figure*}
\begin{figure*}
    \includegraphics[width=0.95\textwidth]{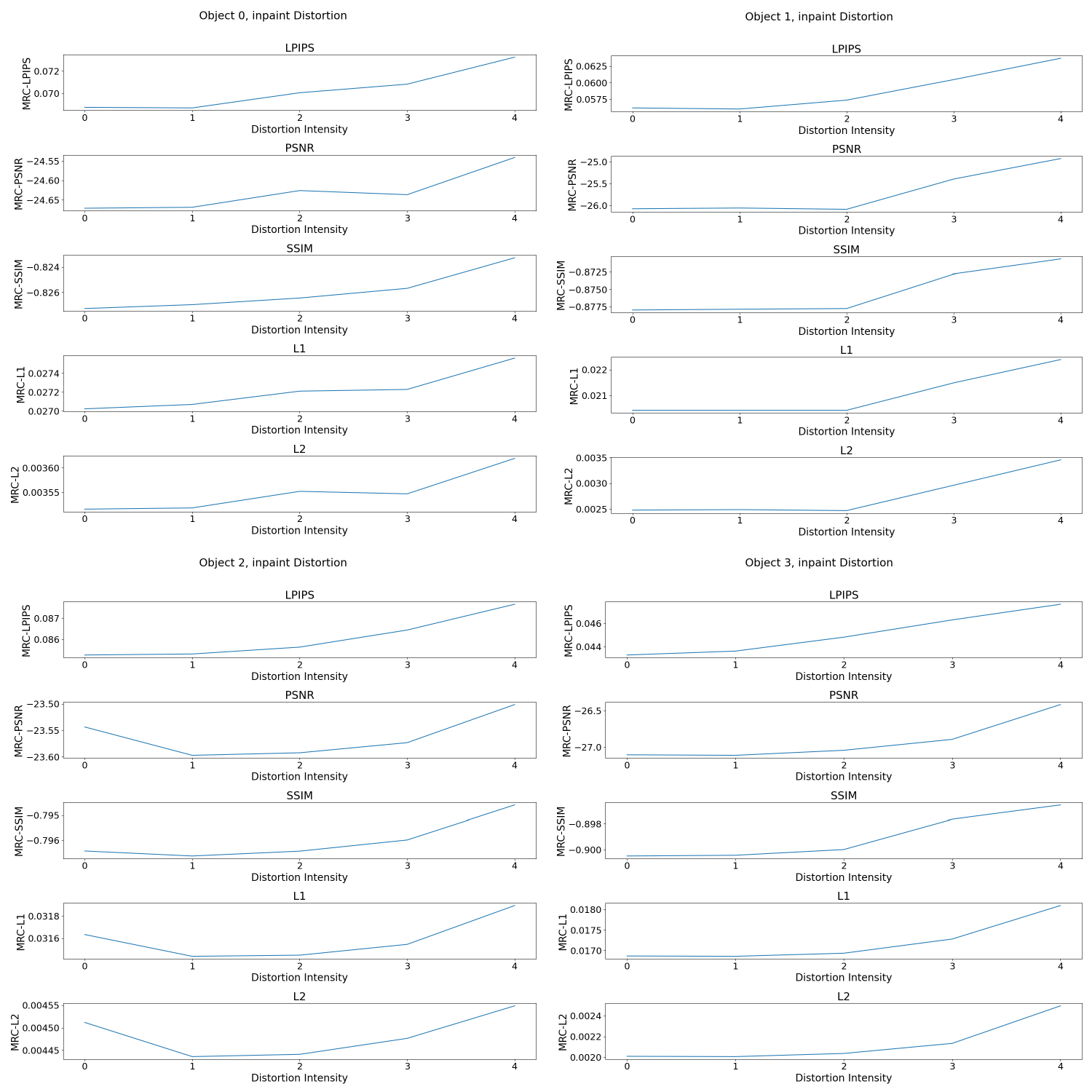}
    \caption{
    Quantitative correlation between five variants of MRC (our default LPIPS, as well as PSNR, SSIM, L1, and L2) and inconsistency introduced by inpaint distortion with increasing intensity on four objects.
    We take negative of the similarity metrics (PSNR and SSIM) for easy comparisons to the distance metrics (LPIPS, L1, and L2).
    LPIPS constantly exhibits monotonically increasing pattern with respect to the increased inconsistency, while other image metrics do not.
    }
    \label{fig:suppl:metricexp:inpaint}
\end{figure*}

\begin{figure*}
    \includegraphics[width=0.95\textwidth]{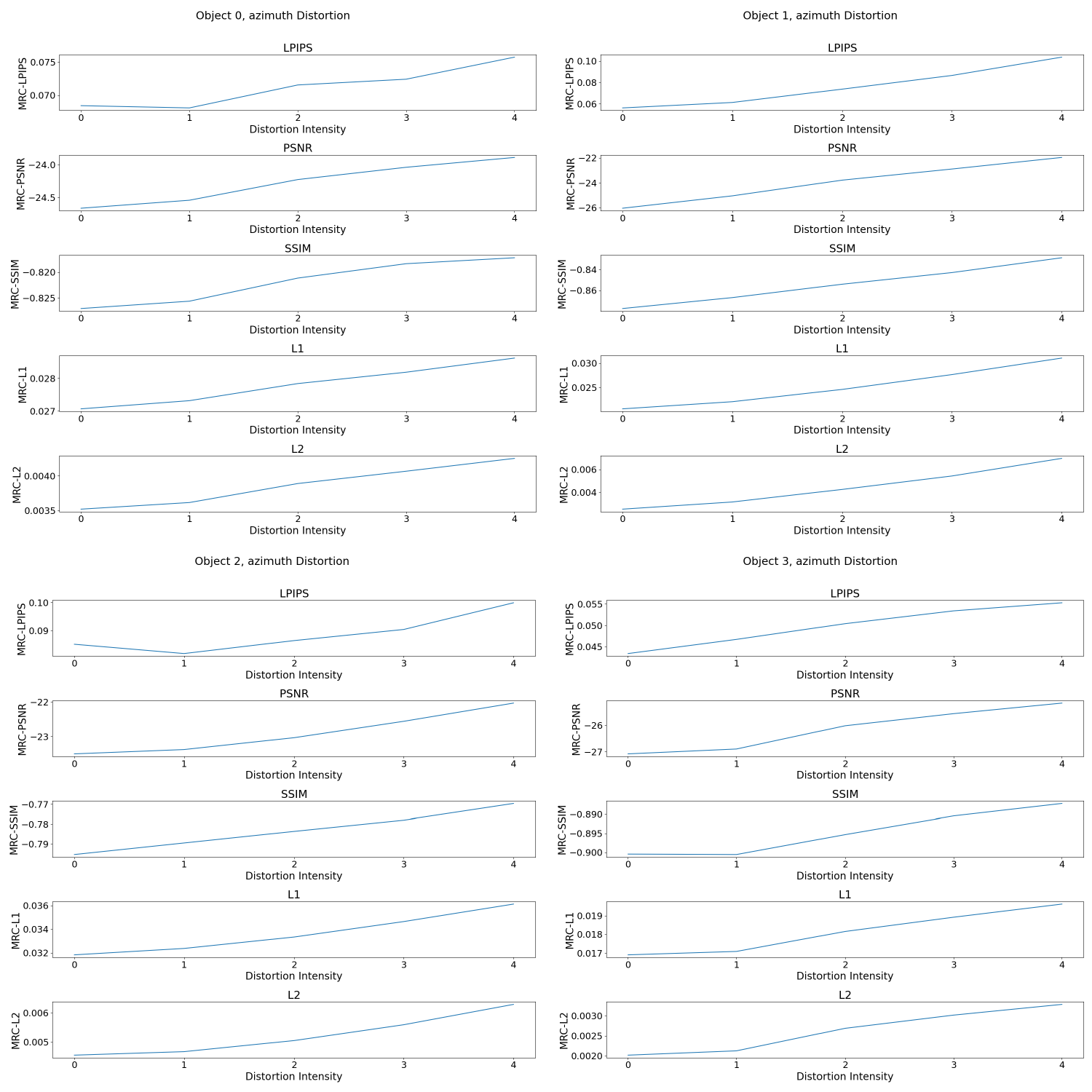}
    \caption{
    Quantitative correlation between five variants of MRC (our default LPIPS, as well as PSNR, SSIM, L1, and L2) and inconsistency introduced by azimuth rotation distortion with increasing intensity on four objects.
    We take negative of the similarity metrics (PSNR and SSIM) for easy comparisons to the distance metrics (LPIPS, L1, and L2).
    All metrics constantly exhibits monotonically, steadily increasing pattern with respect to the increased inconsistency.
    }
    \label{fig:suppl:metricexp:azimuth}
\end{figure*}

\begin{figure*}
    \includegraphics[width=0.95\textwidth]{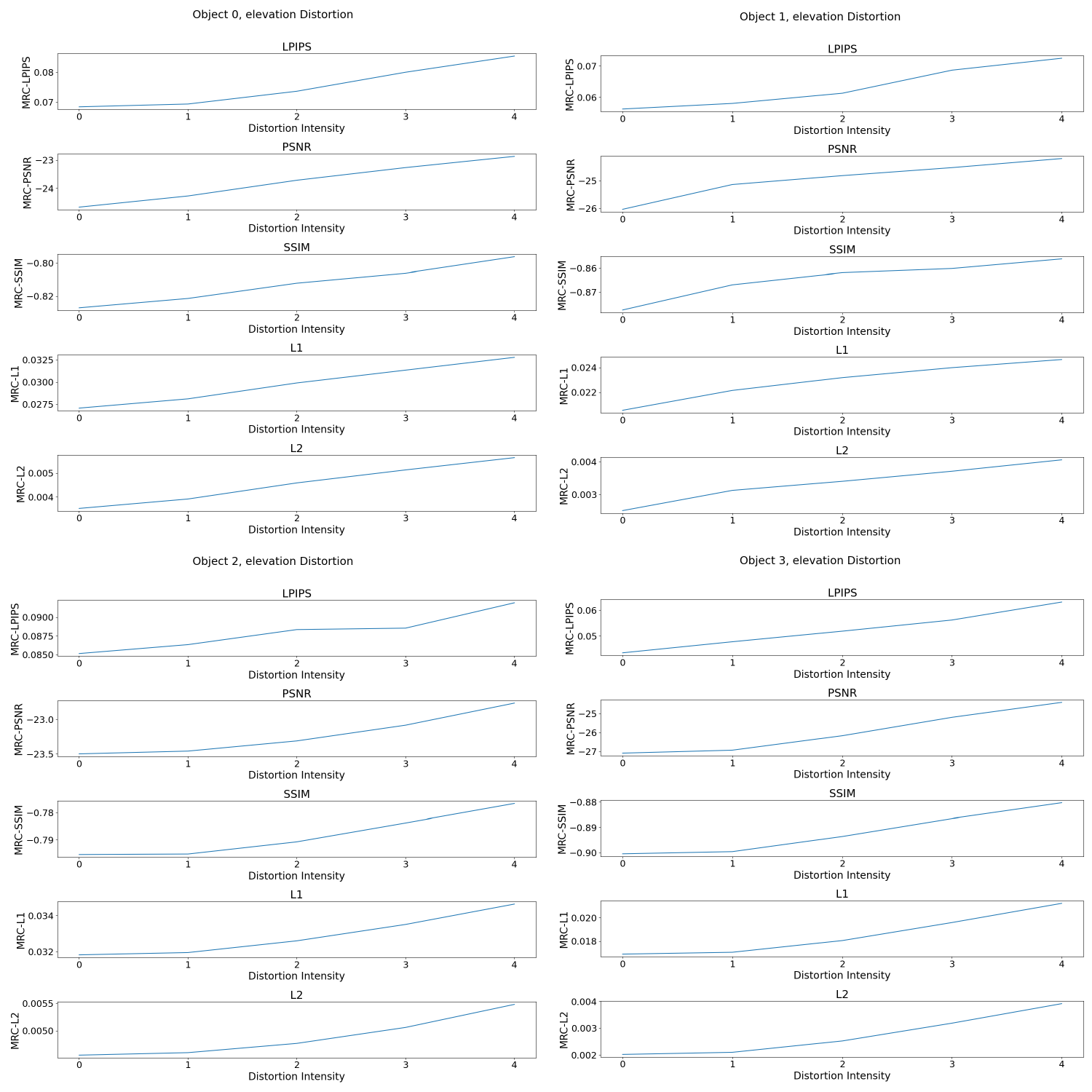}
    \caption{
    Quantitative correlation between five variants of MRC (our default LPIPS, as well as PSNR, SSIM, L1, and L2) and inconsistency introduced by elevation rotation distortion with increasing intensity on four objects.
    We take negative of the similarity metrics (PSNR and SSIM) for easy comparisons to the distance metrics (LPIPS, L1, and L2).
    All metrics constantly exhibits monotonically, steadily increasing pattern with respect to the increased inconsistency.
    }
    \label{fig:suppl:metricexp:elevation}
\end{figure*}

\end{document}